\documentclass{article}
\usepackage{iclr2027_conference,times}

\usepackage{amsmath}
\usepackage{amssymb}
\usepackage[fleqn,tbtags]{mathtools}
\usepackage{multirow}
\usepackage{enumitem}
\usepackage{hyphenat,microtype}
\usepackage{graphicx}
\usepackage{caption,subcaption}
\usepackage{colortbl}
\usepackage{booktabs}
\usepackage{algorithm}
\usepackage{algorithmic}
\usepackage{verbatim}
\usepackage{listings}
\usepackage{wrapfig}
\usepackage{hyperref}
\usepackage{url}
\usepackage[capitalise]{cleveref}

\graphicspath{{./images/}}

\newcommand{\ps}[1]{} 

\newcommand{\tablestyle}[2]{\setlength{\tabcolsep}{#1}
                            \renewcommand{\arraystretch}{#2}
                            \centering
                            \footnotesize}

\definecolor{graycolor}{gray}{.9}

\definecolor{bestblue}{RGB}{181, 204, 228}
\definecolor{secondblue}{RGB}{224, 235, 246}
\newcommand{\bestcell}[1]{\cellcolor{bestblue}#1}
\newcommand{\secondcell}[1]{\cellcolor{secondblue}#1}
\newcommand{\ranknote}{Dark and light blue backgrounds mark the best and
second-best results among generative models, respectively.}

\title{OctMole: Multiscale 3D Molecular Generation with Hierarchical Spatial Flows}

\author{
Yu-Tong Jin$^{1,2}$ \quad
Yi-Yang Tian$^{1,2}$ \quad
Qiwei Ye$^{2}$ \quad
Peng-Shuai Wang$^{1,2}$ \\
$^{1}$Peking University \qquad
$^{2}$Beijing Academy of Artificial Intelligence
}

\iclrfinalcopy 

\hypersetup{pdfkeywords={3D molecular generation, multiscale generation, octree representations, flow matching, shape-conditioned generation}}

\begin{document}

\maketitle
\fancyhead{} 

\begin{abstract}
Generating accurate 3D molecules requires jointly determining a variable
number of atoms, their identities, and geometry across spatial scales: global
shape emerges coarsely, whereas chemical bonds demand fine local precision.
Existing atom-based diffusion and flow models operate largely at a single
scale, while dense spatial fields become costly at high resolution. We
introduce OctMole, a multiscale octree flow framework for 3D molecular
generation. A hierarchical occupancy flow progressively resolves sparse
molecular support from coarse to fine, and a joint atom flow assigns atom types
and continuous in-cell offsets to occupied leaves. This decomposition combines
efficient global structure with subcell geometric precision, while cell
corruption improves robustness to generated support. On QM9 and GEOM-DRUG,
OctMole achieves strong chemical, distributional, and geometric quality, with
particularly clear gains over Uni-3DAR on GEOM-DRUG and substantial
efficiency gains over dense 3D generation. As an additional
capability, intermediate occupancy states and selected regions provide spatial
conditioning without task-specific retraining, supporting multiscale shape
conditioning, regional editing, linker generation, and shape interpolation.
Code and pretrained checkpoints are available
\href{https://github.com/octree-nn/octmole}{here} and
\href{https://huggingface.co/Disembo/octmole}{here}, respectively.
\end{abstract}

\section{Introduction}
\label{sec:intro}

Generating accurate molecular 3D structures is a geometric modeling problem at
a chemically unforgiving scale. A molecule generative model must determine the number of atoms, their chemical identities, and their spatial arrangement.
Even small coordinate errors can alter chemical bonds, valencies, and
conformational energy. These requirements span distinct spatial scales: global
shape and molecular extent emerge at a coarse scale, whereas bond lengths,
angles, and local chemical environments demand fine geometric precision.

Diffusion and flow models have made substantial progress by jointly generating
atomic coordinates and chemical identities~\citep{hoogeboom2022equivariant,
xu2023geometric,Irwin2025SemlaFlow}. Most operate directly on atoms at a single
spatial scale, leaving the coarse-to-fine structure of molecular geometry
implicit. Dense molecular fields provide an explicit spatial representation,
but their cost grows rapidly with resolution~\citep{pinheiro2023voxmol},
creating a tension between global coverage and local accuracy.

Sparse spatial hierarchies offer a natural alternative because molecules occupy only a small fraction of their surrounding 3D domain.
Coarse-to-fine octree sequences and structured 3D latents have demonstrated
the effectiveness of hierarchical representations for 3D shape
generation~\citep{Wei2025OctGPT,Xiang2025TRELLIS}.
Uni-3DAR, the most directly related molecular generator, extends octree
tokenization across several 3D domains~\citep{Uni3DAR2025}. However, it samples
autoregressively and encodes each atom's in-cell coordinates as a set of
24 discrete tokens. The challenge is therefore to combine efficient multiscale
generation with the continuous local precision needed for chemically plausible
geometry.

We introduce \emph{OctMole}, a multiscale octree flow framework for 3D
molecular generation. It first generates spatial support with a
\emph{Hierarchical Occupancy Flow}, then assigns atom types and continuous
coordinates with a \emph{Joint Atom Flow}. The occupancy flow starts from a
single root cell and progressively resolves sparse molecular support from
coarse to fine, jointly
generating all cells at each scale; its occupied leaves determine the number
and approximate locations of atoms. Conditioned on these finest occupied cells, the atom flow jointly generates categorical atom types and continuous in-cell
offsets. The hierarchy captures global molecular structure efficiently, while
the continuous offsets recover chemically sensitive local geometry beyond the
grid resolution. Optional cell corruption makes the atom flow more robust to
errors in generated support.

Across QM9 and GEOM-DRUG, OctMole generates chemically plausible molecules with
strong distributional fidelity and local geometric accuracy. The gains are
particularly clear on the larger and more challenging GEOM-DRUG dataset, where
OctMole achieves 86.82\% molecular stability and 96.90\% connectedness,
representing relative improvements of 11.7\% and 7.8\%, respectively, over
Uni-3DAR, while outperforming it on five of the six distributional distances.
Moreover, OctMole remains the most stable generator across all atom-count bins.
In a system-level comparison, it also provides substantial gains in training and sampling efficiency over dense
3D generation, while our analyses characterize the remaining trade-offs in
diversity and energetic relaxation.

As an additional capability, intermediate occupancy states and selected spatial
regions can condition the same generator without task-specific retraining. We
use this interface for multiscale shape conditioning, regional editing, linker
generation, and shape interpolation, extending the framework to spatially
conditioned generation.
Our contributions are as follows:
\begin{itemize}[leftmargin=*,itemsep=2pt,topsep=3pt]
  \item We introduce a multiscale octree flow framework that models small
  molecules as sparse typed point clouds and resolves their spatial structure
  from coarse to fine.
  \item We formulate atom-level generation as a coarse-to-fine flow-denoising
  problem with continuous in-cell coordinates, combining hierarchical spatial
  organization with chemically precise local geometry; optional cell corruption
  improves robustness to generated spatial support.
  \item We evaluate this representation through molecular quality, geometric
  and energetic analyses, and spatial conditioning,
  together with applications to shape-conditioned generation and regional
  editing. These studies characterize both its utility and its limitations.
\end{itemize}

\section{Related Work}
\label{sec:related}

\paragraph{3D molecular generation.}
Early diffusion approaches established direct generation of three-dimensional
molecules by jointly modeling chemical identity and atomic coordinates. EDM
denoises atom features and coordinates with an equivariant network~\citep{hoogeboom2022equivariant},
and GeoLDM performs diffusion in a learned equivariant latent
space~\citep{xu2023geometric}. Subsequent methods extend this formulation with
additional molecular structure or supervision: MiDi combines discrete graphs
with continuous coordinates in a mixed diffusion process~\citep{vignac2023midi},
while UniGEM connects molecular generation with property
prediction~\citep{feng2024unigem}. These coordinate-based models typically
sample the atom count before generating a fixed-size molecular state, whereas
Jump Diffusion evolves molecular content and dimensionality within the same
process~\citep{Campbell2023Jump}. VoxMol takes a spatial-field view by
generating molecular densities on a dense voxel grid~\citep{pinheiro2023voxmol}.

Flow matching provided an alternative to diffusion by learning transport
vector fields~\citep{Lipman2023Flow}. SemlaFlow applies equivariant flows jointly to
coordinates and chemical attributes, whereas TABASCO uses a non-equivariant
transformer and reconstructs bonds afterward~\citep{Irwin2025SemlaFlow,
Vonessen2025TABASCO}. Zatom-1 extends multimodal flows with auxiliary predictive tasks,
and Branching Flows supports changes in sample size through split and deletion
events~\citep{Morehead2026Zatom,Billera2025Branching}.

These atom- and graph-based methods generally lack explicit multiscale spatial
support. Uni-3DAR autoregressively generates octree structural and atomic
tokens~\citep{Uni3DAR2025}; OctMole instead uses stage-wise occupancy flows over
octree depths, followed by a joint flow for leaf-level atom types and continuous
subcell coordinates. This preserves continuous geometry while exposing
intermediate spatial scales for conditioning.

\paragraph{Sparse and hierarchical 3D generation.}
Octrees have long provided a compact representation of 3D objects.
Octree Transformer generates shapes autoregressively through hierarchically
structured sequences~\citep{Ibing2023,Wei2025OctGPT}, while OctFusion uses octree-based
diffusion for 3D shape generation~\citep{Xiong2024}. TRELLIS generates sparse
structure followed by features in structured 3D latents, supporting several
output representations~\citep{Xiang2025TRELLIS}. These methods establish
important precedents for hierarchical generation and the separation of spatial
support from its associated features. OctMole brings these ideas to molecular
generation through a hierarchical occupancy flow: stages remain sequential
across octree depths, while cells within each stage interact through joint
denoising.

\paragraph{Geometry-conditioned molecular design.}
Geometry has been incorporated through several complementary interfaces:
ShapeMol conditions equivariant diffusion on molecular surfaces, UniGuide
provides a general sampling-guidance framework, and ShEPhERD jointly models
shape, electrostatics, and pharmacophores~\citep{Chen2023ShapeMol,
Ayadi2024UniGuide,Adams2025Shepherd}. DiffLinker instead conditions linker
generation on molecular fragments and their spatial arrangement~\citep{Igashov2022DiffLinker}. OctMole reuses the
same generator with intermediate occupancy states or selected leaf regions,
letting depth and region determine the resolution and extent of spatial
control. We evaluate condition satisfaction and coordinate-frame sensitivity
without assuming exact equivariance.

\section{Method}
\label{sec:method}

\subsection{Atomic Geometry and Our Pipeline}

\begin{wrapfigure}{r}{0.3\linewidth}
  \vspace{-0.7em}
  \centering
  \includegraphics[width=\linewidth]{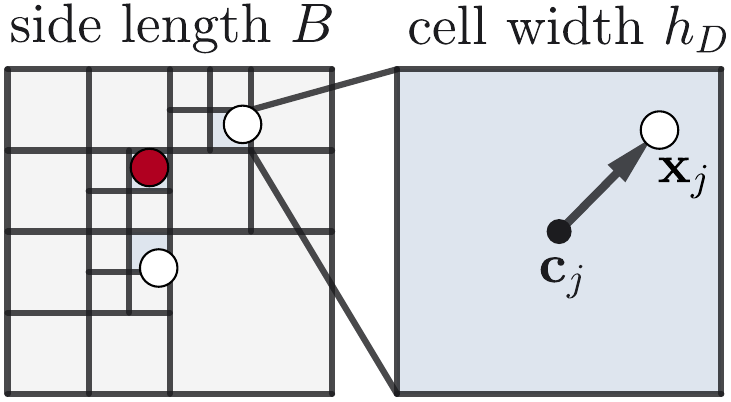}
  \caption{Local spatial representation, illustrated in 2D.
  Left: octree subdivision for a three-atom molecule.
  Right: definition of the local offset within a leaf cell.}
  \label{fig:local_geometry}
  \vspace{-0.8em}
\end{wrapfigure}

We represent a molecule as $M=\{(a_j,\mathbf{x}_j)\}_{j=1}^{N}$, with atom
types $a_j$ and coordinates $\mathbf{x}_j$ in a bounded cube of side length
$B$. Starting from this cube as the root, we recursively split each occupied
cell into eight children and retain those containing atoms until a fixed
depth $D$ is reached. Let $\mathcal{S}_d$ denote the occupied cells at depth
$d$, with width $h_d=B/2^d$. Each level stores $O(N)$ occupied cells, compared
with $8^d$ cells in a full grid; storing $D$ levels requires $O(ND)$ nodes.
\Cref{fig:local_geometry} (left) illustrates this subdivision in 2D for a
three-atom molecule.

At the finest depth $D$, each generated leaf carries one atom. Let
$\mathbf{c}_j$ be the cell center associated with atom $j$. We encode the atom's
displacement from this center in cell-width units using a length $r_j$ and
direction $\mathbf{q}_j$, as shown in \cref{fig:local_geometry} (right). The
local feature $\mathbf{u}_j=(r_j,\mathbf{q}_j)$ then gives the reconstruction
\begin{equation}
  \mathbf{x}_j=\mathbf{c}_j+h_D r_j\mathbf{q}_j.
  \label{eq:molecule_feature}
\end{equation}
Directions are unit vectors for nonzero target displacements. Continuous local
features provide subcell positional precision. The depth $D$ must be large
enough for distinct input atoms to occupy distinct finest cells.

\begin{figure}[t]
  \centering
  \includegraphics[width=\linewidth]{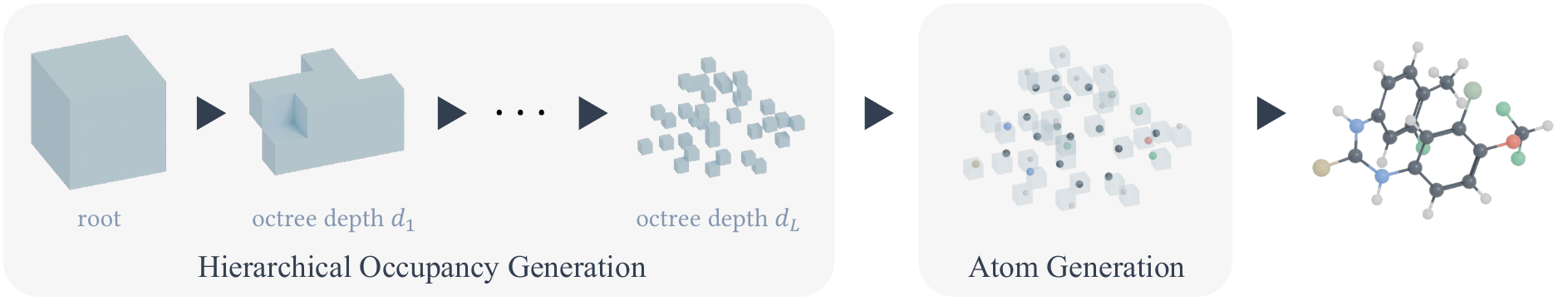}
  \caption{OctMole generates spatial support through hierarchical occupancy
  flows, then jointly generates atom types and continuous local coordinates.
  Intermediate occupancy states provide starting points for spatial conditioning.
  }
  \label{fig:pipeline}
\end{figure}

Our method first determines where atoms can be placed, then
generates their identities and continuous positions. Occupancy is resolved
progressively, allowing each stage to use the spatial context established at
coarser scales.
Let $d_0=0<d_1<\cdots<d_L=D$ be the generation depths, with a fixed increment
$k=d_\ell-d_{\ell-1}$. Starting from the occupied root $\mathcal{S}_0$, we model
\begin{equation}
  p(\mathcal{S}_{d_1},\ldots,\mathcal{S}_D,A,U)
  =\prod_{\ell=1}^{L}p_\theta(\mathcal{S}_{d_\ell}\mid
  \mathcal{S}_{d_{\ell-1}},d_\ell)\,
  p_\phi(A,U\mid\mathcal{S}_D).
  \label{eq:factorization}
\end{equation}
Here $A=(a_j)_{j=1}^{N}$ and $U=(\mathbf{u}_j)_{j=1}^{N}$ collect the atom
types and local features associated with the finest occupied cells. The occupancy model $\theta$
shares parameters across depths; the atom model $\phi$ is trained separately.
\Cref{fig:pipeline} illustrates this decomposition.
We detail the occupancy and atom models in the following two subsections.

\subsection{Hierarchical Occupancy Flow}

Each occupied parent has $8^k$ candidate descendants. We collect their binary
occupancies, mapped to $\{-1,1\}$, denoted as $Y$. A sparse Diffusion
Transformer (DiT) jointly denoises
all parent tokens, conditioned on their spatial locations, the target depth,
and flow time; architecture details are provided in
\cref{app:network_architecture}. Following flow
matching~\citep{Lipman2023Flow}, we interpolate between
Gaussian noise $E$ and the target $Y$ and predict the clean endpoint~\citep{Li2025}:
\begin{equation}
  Y_t=(1-t)E+tY,\qquad
  v_\theta=\frac{\widehat{Y}_\theta(Y_t,t)-Y_t}{\max(1-t,\varepsilon)},\qquad
  \mathcal{L}_{\mathrm{occ}}
  =\mathbb{E}\!\left[w(t)\|\widehat{Y}_\theta-Y\|^2\right].
  \label{eq:flow_loss}
\end{equation}
Conditioning arguments are suppressed for readability. We use
$w(t)=\max(1-t,\varepsilon)^{-2}$, where $\varepsilon>0$ bounds the weight
near $t=1$.
During training, we sample a growth depth and the corresponding ground-truth
parent support. At inference, we integrate the flow and threshold its endpoint
to determine the next set of occupied cells. All parents are processed jointly
within a stage; stages follow the hierarchy sequentially.

\subsection{Joint Atom Flow}

Conditioned on the finest occupied cells $\mathcal{S}_D$, a second sparse DiT
jointly processes atom types and local features; architecture details are
given in \cref{app:network_architecture}. Continuous features follow
$U_t=(1-t)E_U+tU$, where $E_U$ is Gaussian noise, and its loss function
$\mathcal{L}_{\mathrm{coord}}$ uses the continuous endpoint objective in \cref{eq:flow_loss}.
For atom types, a uniform reference
distribution $p_{\mathrm{ref}}$ over the chemical elements in the dataset defines the discrete path
\begin{align}
  q_t(a_{j,t}\mid a_j)
  &= (1-t)p_{\mathrm{ref}}(a_{j,t}) + t\delta_{a_j}(a_{j,t}).
  \label{eq:type_path}
\end{align}
Let $A_t=(a_{j,t})_{j=1}^{N}$. The atom type loss is averaged over atoms,
\begin{align}
  \mathcal{L}_{\mathrm{type}}
  &= \mathbb{E}\!\left[-\frac{1}{N}\sum_j
  \log p_\phi(a_j\mid A_t,U_t,\mathcal{S}_D,t)\right].
  \label{eq:type_loss}
\end{align}

The complete loss function for the joint atom flow is
$\mathcal{L}_{\mathrm{atom}}=\mathcal{L}_{\mathrm{coord}}+\lambda\mathcal{L}_{\mathrm{type}}$.
Here $\lambda$ balances the coordinate and atom-type objectives.
Separate output heads share the same molecular context and predict clean local
features and endpoint type probabilities. Sampling advances the
continuous features together with discrete type transitions induced by these
probabilities~\citep{Campbell2024}. The resulting atom count is $N=|\mathcal{S}_D|$.
Bonds are inferred from the generated types and coordinates during evaluation.

\paragraph{Cell corruption.}
The atom model is trained on ground-truth cells but receives generated cells
at inference. To reduce sensitivity to this exposure bias, cell corruption
shifts selected training cells while compensating their local offsets to
preserve the target geometry; details appear in \cref{app:cell_corruption}.

\subsection{Spatial Conditioning}
For a reference molecule, we extract its occupied cells at a chosen depth
$d_m$ and resume the remaining occupancy transitions in
\cref{eq:factorization}, followed by the atom flow. Finer starting states leave
fewer spatial decisions; at depth $D$, only atom types and local features are
sampled. For regional editing, we retain selected finest cells, regrow the
remaining support from coarse cells, and combine both supports before atom
generation. Because these conditions fix spatial support rather than atom
identities, they provide resolution- and region-based control without
task-specific models, but do not guarantee preserved fragments or exact final
geometry. We evaluate these properties empirically, including coordinate-frame
sensitivity.

\section{Experiments}
\label{sec:result}

We evaluate unconditional molecular generation, continuous geometric quality,
and the spatial control exposed by intermediate octree states.

\subsection{Experimental Setup}

\paragraph{Datasets and models.}
We train separate models on QM9~\citep{ramakrishnan2014quantum} and
GEOM-DRUG~\citep{axelrod2022geom}, covering small organic molecules and larger,
flexible drug-like molecules. Unless stated otherwise, all experiments use
maximum octree depth $D=6$, depth increment $k=2$, and 50-step Heun sampling.
Main results use cell corruption on GEOM-DRUG and no corruption on QM9.
Dataset, architecture, training, and sampling details are provided in
\cref{app:sec:datasets,app:sec:implementation}.

\paragraph{Baselines and sample selection.}
We compare with EDM~\citep{hoogeboom2022equivariant},
GeoLDM~\citep{xu2023geometric}, UniGEM~\citep{feng2024unigem}, and
Uni-3DAR~\citep{Uni3DAR2025}; UniGEM uses additional molecular properties
during training. All unconditional results use 10,000 final samples per model
under a shared chemical assessment. OctMole retains approximately 10\% of its
QM9 candidates and 20\% of its GEOM-DRUG candidates by mean atom-type log
probability, as does Uni-3DAR under its native protocol; the other baselines do
not rerank.

\paragraph{Metrics.}
We infer bonds with Open Babel~\citep{oboyle2011openbabel} and perform chemical
checks with RDKit. Molecule stability requires every atom to have an allowed
valency; validity measures successful sanitization of the largest fragment;
connectedness is measured before fragment selection; and uniqueness is the
fraction of distinct canonical SMILES among valid samples. We compare
molecular size, atom and bond types, valencies, bond lengths, and bond angles
against reference distributions using total variation or Wasserstein-1
distance. Exact definitions are given in \cref{app:sec:evaluation}.

\subsection{De Novo 3D Molecule Generation}
\label{sec:molecule_generation}

\Cref{tab:openbabel_quality} compares chemical quality and diversity. On
GEOM-DRUG, OctMole achieves 86.82\% molecule stability, 98.20\% validity, and
96.90\% connectedness, compared with 77.76\%, 96.49\%, and 89.89\% for
Uni-3DAR. Metrics on QM9 are already near saturation. OctMole achieves 99.23\% molecule stability and 99.72\%
validity, but its 80.77\% uniqueness is substantially below the baselines'
95.84--98.49\%. We suspect that memorization on the relatively small training set contributes to repetition,
with confidence-based selection further concentrating the outputs.

\begin{table*}[t]
  \caption{Chemical quality of unconditional 3D molecule generation under our
  Open Babel evaluation. All values are percentages and higher is better.
  UniGEM$^{*}$ uses additional molecular property information. All models are
  evaluated on 10,000 final samples. \ranknote}
  \label{tab:openbabel_quality}
  \tablestyle{6pt}{1.12}
  \begin{tabular}{lccccc}
    \toprule
    \textbf{Method} & \textbf{Atom Sta.} & \textbf{Mol. Sta.} &
    \textbf{Valid.} & \textbf{Unique} & \textbf{Connected} \\
    \midrule
    \multicolumn{6}{c}{\textbf{QM9}} \\
    \midrule
    Data & 99.900 & 99.300 & 99.500 & -- & -- \\
    EDM~\citep{hoogeboom2022equivariant} & 99.841 & 97.230 & 98.740 & \bestcell{98.491} & 99.630 \\
    GeoLDM~\citep{xu2023geometric} & 99.909 & 98.430 & 99.090 & \secondcell{98.032} & \secondcell{99.700} \\
    UniGEM$^{*}$~\citep{feng2024unigem} & 99.815 & 98.550 & 99.500 & 96.010 & 99.350 \\
    Uni-3DAR~\citep{Uni3DAR2025} & \bestcell{99.976} & \bestcell{99.620} & \bestcell{99.800} & 95.842 & 99.680 \\
    OctMole & \secondcell{99.950} & \secondcell{99.230} & \secondcell{99.720} & 80.766 & \bestcell{99.810} \\
    \midrule
    \multicolumn{6}{c}{\textbf{GEOM-DRUG}} \\
    \midrule
    Data & 99.800 & 92.300 & 98.400 & -- & -- \\
    EDM~\citep{hoogeboom2022equivariant} & 97.950 & 43.340 & 87.920 & \bestcell{100.000} & 42.320 \\
    GeoLDM~\citep{xu2023geometric} & 98.670 & 58.330 & 90.140 & \secondcell{99.990} & 51.580 \\
    UniGEM$^{*}$~\citep{feng2024unigem} & 98.470 & 54.310 & 88.880 & 99.930 & 43.160 \\
    Uni-3DAR~\citep{Uni3DAR2025} & \secondcell{99.011} & \secondcell{77.760} & \secondcell{96.490} & 98.031 & \secondcell{89.890} \\
    OctMole & \bestcell{99.624} & \bestcell{86.820} & \bestcell{98.200} & 99.969 & \bestcell{96.900} \\
    \bottomrule
  \end{tabular}
\end{table*}

\begin{figure}[t]
  \centering
  \includegraphics[width=0.65\linewidth,trim=0 3.5cm 0 0,clip]{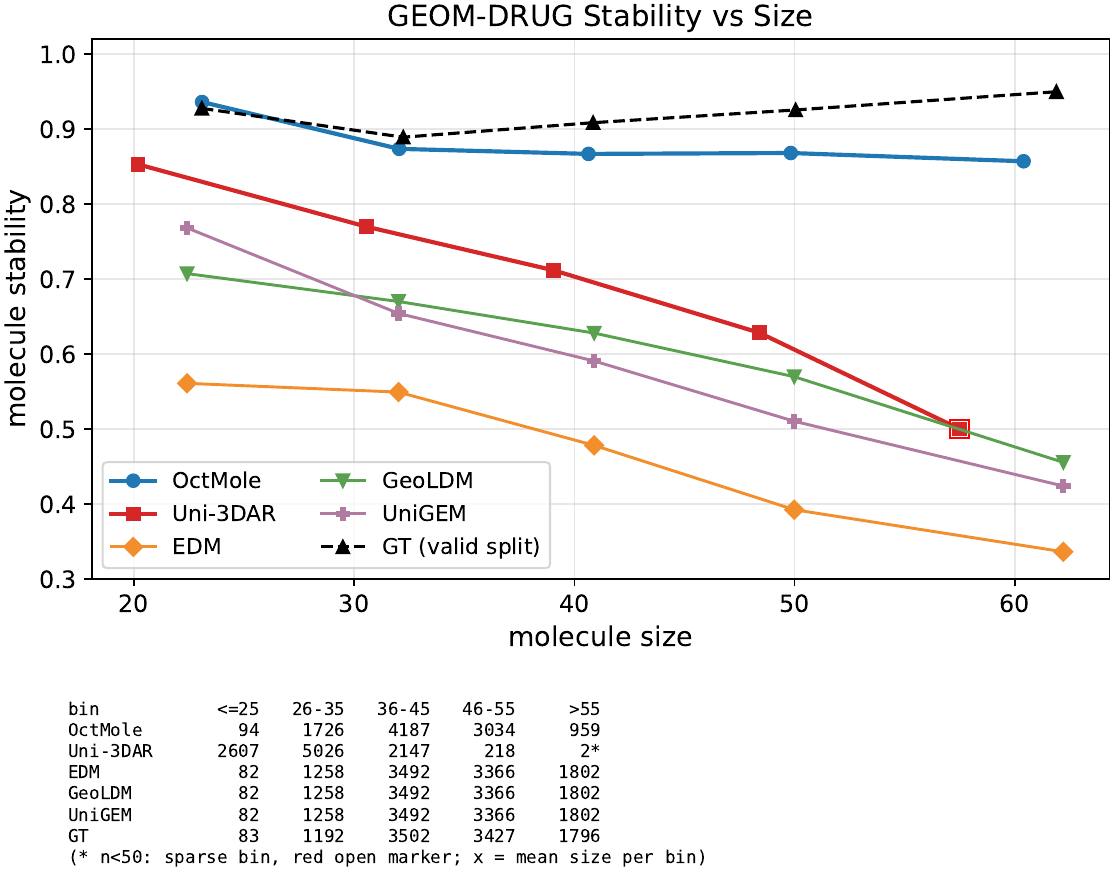}
  \caption{Open Babel molecule stability by atom-count bin on 10,000
  GEOM-DRUG molecules per group. Marker positions are the mean atom count
  within each bin. OctMole is the most stable generator in every bin and degrades only
  mildly with size. Uni-3DAR has only two samples in the largest bin (red open marker),
  so its value there is not statistically informative.}
  \label{fig:stability_vs_size}
\end{figure}

OctMole's advantage is not explained by generating easier, smaller molecules.
As shown in \cref{fig:stability_vs_size}, it has the highest GEOM-DRUG
molecule stability in every size bin and degrades only mildly as molecule size
increases. Across the well-sampled bins up to 55 atoms, its margin over
Uni-3DAR widens from 8.3 to 24.0 percentage points.

\subsection{Continuous Geometry and Physical Quality}
\label{sec:geometry}

Across the six distributional distances, OctMole outperforms Uni-3DAR on
molecular size, atom type, bond type, valency, and bond angle on GEOM-DRUG;
Uni-3DAR has the lower bond-length distance. The complete comparison and a
size-stratified analysis are given in
\cref{app:tab:openbabel_distribution,app:fig:distribution_vs_size}.

We further optimize valid, connected molecules with GFN2-xTB. End-to-end yield
is the fraction of all 10,000 final samples that are valid, connected, and
converge under optimization; energy and RMSD are medians over converged
molecules. \Cref{tab:xtb_compact} shows that OctMole attains the highest yield
on both datasets. Its QM9 median relaxation RMSD of 0.0628\,\AA{} is close to
the data value of 0.0574\,\AA{}. On GEOM-DRUG, Uni-3DAR has lower median
relaxation energy and a similar RMSD, but preferentially generates smaller
molecules. Complete xTB results and size-conditioned diagnostics appear in
\cref{app:xtb_evaluation,app:size_diagnostics}.

\begin{table}[t]
  \caption{GFN2-xTB comparison. Yield is relative to all 10,000 final
  samples; energy (kcal/mol/atom) and RMSD (\AA) are medians over converged
  valid, connected molecules; atoms is the mean count in the corresponding
  xTB-evaluated sample set.}
  \label{tab:xtb_compact}
  \tablestyle{5pt}{1.1}
  \begin{tabular}{llrrrr}
    \toprule
    Dataset & Method & Yield (\%) & Energy & RMSD & Atoms \\
    \midrule
    QM9 & Data & -- & 0.0690 & 0.0574 & 18.0 \\
        & Uni-3DAR & 99.47 & 0.2178 & 0.2349 & 17.2 \\
        & OctMole & 99.53 & \textbf{0.1149} & \textbf{0.0628} & \textbf{18.4} \\
    \midrule
    GEOM-DRUG & Data & -- & $\approx0$ & $\approx0$ & 46.6 \\
        & Uni-3DAR & 86.51 & \textbf{0.2988} & 0.6019 & 30.1 \\
        & OctMole & \textbf{94.86} & 0.6335 & 0.6067 & \textbf{43.7} \\
    \bottomrule
  \end{tabular}
\end{table}

\subsection{Multiscale Spatial Conditioning}
\label{sec:multiscale_control}

\begin{figure}[t]
  \centering
  \includegraphics[width=\linewidth]{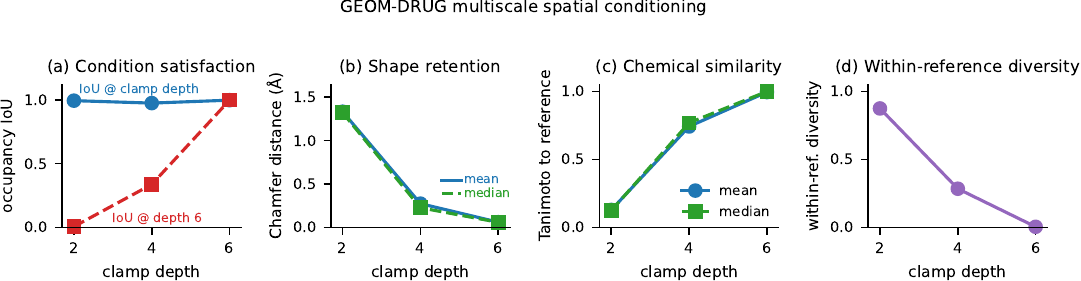}
  \caption{Multiscale spatial conditioning on GEOM-DRUG. Coarse conditions
  leave substantial chemical freedom, intermediate conditions retain shape
  while supporting diverse chemistry, and finest conditions approach
  reconstruction.}
  \label{fig:multiscale_control}
\end{figure}

We fix reference occupancy at depths 2, 4, and 6 and reuse the same generator
to sample the remaining structure. We use 256 references per
dataset and retain 2,000 final samples per depth from shared candidate
budgets. IoU measures satisfaction of the supplied occupancy; Chamfer distance
is computed in the shared frame; diversity is one minus within-reference
pairwise Tanimoto similarity. Joint success requires a sample to be valid and
stable, within 1\,\AA{} Chamfer of the reference, and chemically distinct from
it. \Cref{tab:multiscale_control_main} summarizes the control--diversity
trade-off. \Cref{fig:multiscale_control} illustrates the corresponding transition from
coarse shape guidance to near-reconstruction on GEOM-DRUG.

\begin{table*}[t]
  \caption{Multiscale spatial conditioning. IoU@$d$ is occupancy overlap at
  the clamped depth; Joint is the percentage of final samples satisfying
  chemical quality, shape, and novelty simultaneously. Full chemical
  statistics and confidence intervals are in
  \cref{app:tab:multiscale_control_full}.}
  \label{tab:multiscale_control_main}
  \tablestyle{7pt}{1.1}
  \begin{tabular}{llrrrr}
    \toprule
    Dataset & Depth & IoU@$d$ & Chamfer (\AA) & Diversity & Joint (\%) \\
    \midrule
    QM9 & 2 & 1.00 & 1.00 & 0.88 & \textbf{41.3} \\
        & 4 & 0.97 & 0.08 & 0.01 & 4.2 \\
        & 6 & 0.99 & 0.03 & 0.00 & 0.0 \\
    \midrule
    GEOM-DRUG & 2 & 0.99 & 1.33 & 0.88 & 1.1 \\
              & 4 & 0.98 & 0.27 & 0.28 & \textbf{53.3} \\
              & 6 & 1.00 & 0.06 & 0.00 & 1.7 \\
    \bottomrule
  \end{tabular}
\end{table*}

The supplied occupancy is recovered almost exactly at every depth, with IoU
between 0.97 and 1.00. The conditioning depth instead determines how much
chemical freedom remains. On GEOM-DRUG, depth 4 best balances shape retention
and exploration, reaching 53.3\% joint success; on the smaller QM9 molecules,
the coarser depth-2 condition performs best at 41.3\%. Finest-depth conditions
approach reconstruction and therefore provide little diversity. The complete
metrics, confidence intervals, reference-coverage analysis, and coordinate
sensitivity tests are provided in
\cref{app:multiscale_control,app:coordinate_sensitivity}. In particular, our
model is robust to grid-origin shifts, but depth-4 GEOM-DRUG conditioning remains
sensitive to rotations despite near-exact discrete condition satisfaction.

\section{Ablations}
\label{sec:ablations}

\paragraph{Octree versus dense generation.}
\label{sec:sparse_ablation}
We replaced the hierarchical occupancy flow with flow matching on a dense $64^3$ grid,
using a DiT trained with the same number of epochs and GPUs.
\Cref{tab:sparse_compact} summarizes this system-level comparison. The octree
occupancy model reduces observed training time by at least $5\times$ and
sampling time by at least $4\times$ on both datasets. It also achieves
significantly better molecule stability.
At higher resolutions, dense grids grow cubically while octree token counts grow
roughly linearly, demonstrating the efficiency of our method. Full configurations,
quality metrics, and resolution scaling are provided in \cref{app:sparse_details}.

\paragraph{Cell corruption.}
\label{sec:corruption}
Cell corruption targets the support mismatch between atom-flow training and
inference. Under generated support, it raises GEOM-DRUG molecule stability
from 71.15\% to 86.82\% and connectedness from 90.60\% to 96.90\%
(\cref{tab:corruption_compact}). The effect is weaker on QM9, where scores are
near saturation. Separate GT-support tests show negligible changes, indicating
that corruption primarily improves robustness to occupancy errors rather than
clean support.

\begin{table*}[t]
  \begin{minipage}[t]{0.47\linewidth}
    \vspace{0pt}
    \centering
    \captionof{table}{System-level efficiency comparison. Unparenthesized and
    parenthesized molecule stabilities are Dense and Octree, respectively.}
    \label{tab:sparse_compact}
    \tablestyle{3pt}{1.05}\scriptsize
    \begin{tabular}{lrrr}
      \toprule
      Dataset & Train speedup & Sample speedup & Mol. Sta. (\%) \\
      \midrule
      QM9 & $6.85\times$ & $5.51\times$ & 69.50 (\textbf{99.90}) \\
      GEOM-DRUG & $5.13\times$ & $4.35\times$ & 4.91 (\textbf{84.40}) \\
      \bottomrule
    \end{tabular}
  \end{minipage}
  \hfill
  \begin{minipage}[t]{0.50\linewidth}
    \vspace{0pt}
    \centering
    \captionof{table}{Cell corruption ablation on 10,000 samples.
    Unparenthesized and parenthesized values are without and with
    cell corruption, respectively.}
    \label{tab:corruption_compact}
    \tablestyle{3pt}{1.05}\scriptsize
    \begin{tabular}{lcrr}
      \toprule
      Dataset & Mol. Sta. (\%) & Connected (\%) \\
      \midrule
      QM9       & 99.23 (\textbf{99.73}) & 99.81 (\textbf{99.96}) \\
      GEOM-DRUG & 71.15 (\textbf{86.82}) & 90.60 (\textbf{96.90}) \\
      \bottomrule
    \end{tabular}
  \end{minipage}
\end{table*}

\paragraph{Continuous offsets and quantization.}
\label{sec:offset_ablation}
Snapping OctMole-generated atoms to cell centers while fixing support and atom types
increases GEOM-DRUG bond-length $W_1$ from 0.017 to 0.135\,\AA{} and
bond-angle $W_1$ from 1.12 to 7.97 degrees. Directly quantizing 300 validation conformers
shows the same grid effect (\cref{fig:quantization_xtb}): at $64^3$ resolution,
quantization alone induces a median GFN2-xTB relaxation of
20.9\,kcal/mol/atom, whereas OctMole with learned continuous offsets yields
0.63\,kcal/mol/atom. The complete bond-geometry diagnostic is provided in
\cref{app:offset_geometry}.

\paragraph{Hierarchy settings.}
\label{sec:hierarchy_ablation}
Let $D$ be the maximum octree depth which determines the final $2^D$ grid
resolution along each axis, and $k$ be the depth increment between
successive occupancy-generation stages. At $D=6$,
$k=2$ reduces the schedule from six stages to three while retaining quality
comparable to or better than $k=1$. Increasing to $D=8$, however, substantially
degrades quality. We therefore use $D=6,k=2$ as our default setting, corresponding
to depth generation at $d=2,4,6$ and a final resolution of $64^3$.
The complete comparison is reported in \cref{app:tab:hierarchy_ablation}.

\paragraph{Sampling steps and candidate selection.}
\label{sec:sampling_ablation}
We uses a 50-step Heun integration by default.
Reducing 50 steps to 25 lowers generation time
by $1.77\times$ on QM9 and $1.89\times$ on GEOM-DRUG ($1.2\times$ and $1.3\times$ faster than Uni-3DAR, respectively),
with a drop of at most 1.80\% in molecule stability; 10 steps degrades stability and connectedness.
Confidence reranking improves chemical quality but changes the output
distribution. Full sampling-step and selection analyses appear in
\cref{app:sampling_selection}.

\section{Applications}
\label{sec:applications}

All applications reuse the same trained models without task-specific
fine-tuning. Conditions act only on spatial support; atom types and coordinates
are always generated.

\paragraph{Shape-conditioned ligand reconstruction and exploration.}
\label{sec:app_shape}

\begin{figure*}[t]
  \centering
  \begin{minipage}[t]{0.5\linewidth}
    \vspace{0pt}
    \centering
    \includegraphics[width=\linewidth]{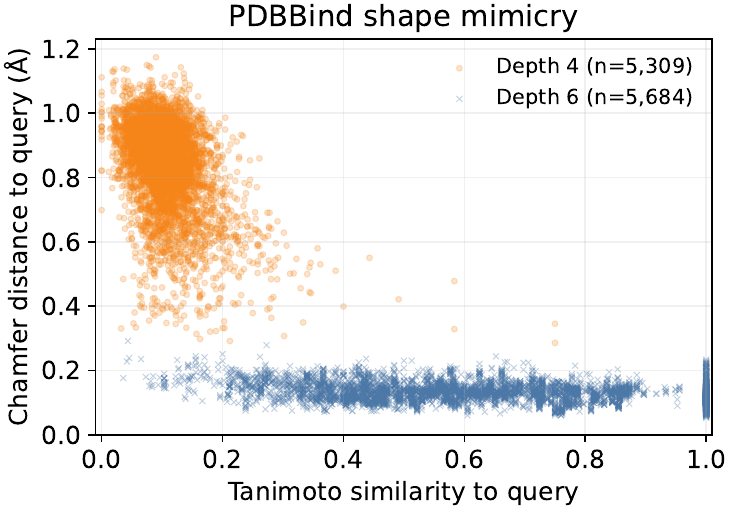}
    \captionof{figure}{Shape-conditioned PDBBind generation over 354 ligands
    with 20 candidates per condition. Points compare Chamfer distance and
    Tanimoto similarity; orange circles and blue crosses denote depth-4 and
    depth-6 conditioning, respectively.}
    \label{fig:shape_mimicry_scatter}
  \end{minipage}
  \hfill
  \begin{minipage}[t]{0.43\linewidth}
    \vspace{0pt}
    \centering
    \includegraphics[width=\linewidth]{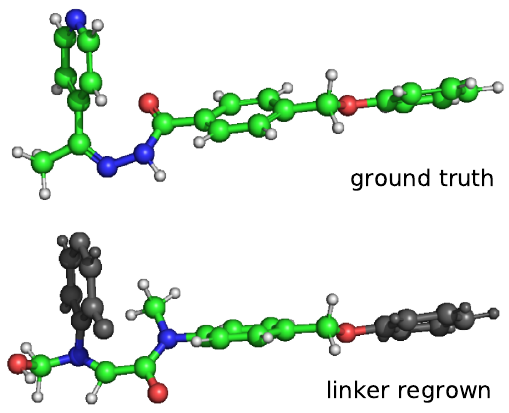}
    \captionof{figure}{A linker-generation case. End-fragment occupancy is
    fixed in dark gray, and the regenerated linker is shown in element
    colors.}
    \label{fig:linker_case}
  \end{minipage}
\end{figure*}

We condition on the occupancy of 354 eligible PDBBind ligands and generate 20
candidates per ligand. Finest depth-6 occupancy nearly determines the ligand:
32.2\% of queries have a candidate with Morgan-fingerprint Tanimoto similarity
$\geq0.99$. Coarser depth-4 occupancy instead retains shape while exploring
chemistry: mean Chamfer distance is 0.840\,\AA{}, mean Tanimoto similarity is
0.107, and 69.6\% of candidates are valid, within 1\,\AA{} of the query, and
have similarity at most 0.3; 297 of 354 queries yield at least one such mimic.
\Cref{tab:shape_conditioning,fig:shape_mimicry_scatter} summarize this
trade-off. No pocket information is supplied, so shape matching does not imply
preserved binding; full details are in \cref{app:shape_application}.

\begin{table}[t]
  \centering
  \caption{Shape retention, chemical similarity, and molecular quality under
  coarse and fine PDBBind ligand occupancy conditions. IoU@4 measures
  generated--query occupancy overlap at depth 4.}
  \label{tab:shape_conditioning}
  \tablestyle{6pt}{1.12}
  \begin{tabular}{lccccc}
    \toprule
    Condition & IoU@4 & Chamfer (\AA) & Tanimoto & Valid & Stable \\
    \midrule
    Depth 4 (coarse) & 0.909 & 0.840 & 0.107 & 0.895 & 0.321 \\
    Depth 6 (fine) & 0.968 & 0.131 & 0.663 & 0.954 & 0.831 \\
    \bottomrule
  \end{tabular}
\end{table}

\paragraph{Regional editing and linker generation.}
\label{sec:app_editing}
\label{sec:app_linker}
Retaining one spatial region while regrowing another, or fixing two end
regions while generating a linker, demonstrates that the spatial interface
extends beyond global shape conditioning. Both tasks also reveal that joining
independently generated supports remains a principal failure mode.
\Cref{fig:linker_case} demonstrates a linker generation case.
Full protocols, quantitative results, and examples are provided in
\cref{app:regional_editing,app:linker_generation}.

\paragraph{Shape interpolation.}
\label{sec:app_morph}
Interpolating depth-4 occupancy fields yields consistent shape transitions
across 20 reference pairs: 59 of 60 selected intermediate outputs are valid
and 53 are molecule-stable. The construction and visualizations are provided in
\cref{app:shape_interpolation}.

\section{Conclusion}
\label{sec:conclusion}

\paragraph{Main conclusions.}
We introduced OctMole, a multiscale framework that first generates sparse
octree occupancy from coarse to fine and then jointly generates atom types and
continuous in-cell coordinates. On GEOM-DRUG, OctMole achieves 86.82\%
molecule stability and 96.90\% connectedness, outperforms Uni-3DAR on five of
six distributional distances, remains the most stable generator across
atom-count bins, and attains a 94.86\% end-to-end xTB yield. On
QM9, its median relaxation RMSD of 0.0628\,\AA{} is close to the 0.0574\,\AA{}
data reference.
The same hierarchy also exposes spatial control without
task-specific retraining: supplied occupancies reach 0.97--1.00 IoU, and the
joint shape-and-chemistry success rate peaks at 53.3\% for depth-4 GEOM-DRUG
conditioning and 41.3\% for depth-2 QM9 conditioning. Regional editing,
linker generation, and shape interpolation further demonstrate this interface.

\paragraph{Limitations and scope.}
Several limitations qualify these results.
First, QM9 uniqueness is only
80.77\%, and the remaining relationship between repetition and training-set
memorization requires molecule-level analysis.
Second, octree conditions are tied to a coordinate frame: on GEOM-DRUG, depth-4 molecule
stability falls from 95.40\% to 82.95\% under rotation and to 80.05\% when
rotation and grid-origin shifts are combined.
Third, OctMole does not generate bonds explicitly; bonds are inferred from
atom types and coordinates using Open Babel and RDKit, making chemical metrics
dependent on this inference pipeline.
Finally, xTB energy and RMSD statistics include only valid,
connected samples whose optimizations converge. Reporting end-to-end yield
exposes the size of this subset but does not remove survivor bias within it.

\paragraph{Future directions.}
Promising directions include diversity-aware generation and selection,
reranking-free evaluation, rotation- and grid-robust hierarchical
representations, explicit joint modeling of bond structure, and physical
assessment protocols that better account for failed or filtered samples.

\subsection*{AI use statement}
In this work, we used generative AI tools to design and provide feedback on
research methodology and experiments, implement methods, assist with
translation, and interpret results. We did not use generative AI tools to
generate synthetic datasets, develop theoretical models or conceptual frameworks,
clean or reformat datasets, or support qualitative and thematic analysis;
mathematical-claim and proof-related uses were not applicable to this work.
Additionally, we used generative AI tools to create or modify scientific
figures, create and edit software code, draft and edit parts of the manuscript,
summarize and analyze existing literature, brainstorm, source information,
identify relevant literature, format references, and propose the title or
keywords.

We reviewed all AI-assisted work. The authors manually checked every citation
against its original source, reviewed AI-assisted code and executed the
relevant pipelines, and cross-checked figures and numerical claims against the
underlying experimental records. Methodological suggestions were evaluated
through experiments, and only claims supported by the reported results were
retained. AI-assisted prose and translations were edited by the authors for
technical accuracy, clarity, and consistency with the rest of the manuscript.
We take responsibility for the final content of this work, including text,
claims, code, figures, and other artifacts produced with the aid of generative
AI.

\subsection*{Ethics statement}
This work focuses on methodological research for molecular generation using publicly
available benchmarks and datasets and does not involve human subjects or personally identifiable
information. Generative models for molecules may have both beneficial scientific applications
and potential misuse in the design of harmful compounds. Our experiments are limited to
standard research benchmarks and do not evaluate or optimize for harmful biological or chemical properties.

\subsection*{Reproducibility statement}
Our code is available at \url{https://github.com/octree-nn/octmole}, and pretrained
checkpoints are available at \url{https://huggingface.co/Disembo/octmole}.
Implementation details are provided in \cref{app:sec:implementation}.

\bibliography{src/ref/reference}
\bibliographystyle{iclr2027_conference}

\clearpage
\appendix
{
\hypersetup{linkcolor=black}
\begin{center}
  {\LARGE\bfseries Appendix\par}
  \vspace{1.5em}
  {\large\bfseries Contents\par}
\end{center}
\vspace{1em}
\makeatletter
\setcounter{tocdepth}{2}
\@starttoc{atc}
\makeatother
}
\clearpage
\makeatletter
\renewcommand{\addcontentsline}[3]{%
  \def\octmoletoc{toc}%
  \def\octmoletarget{#1}%
  \ifx\octmoletoc\octmoletarget
    \addtocontents{atc}{\protect\contentsline{#2}{#3}{\thepage}{\@currentHref}%
      \protected@file@percent}%
  \fi}
\makeatother
\section{Datasets}
\label{sec:appendix}
\label{app:sec:datasets}

We follow the preprocessing and data splits of
EDM~\citep{hoogeboom2022equivariant} for both benchmarks.
\paragraph{QM9.}
QM9~\citep{ramakrishnan2014quantum} contains small organic molecules with up
to nine heavy atoms. Starting from approximately 134K molecules, the EDM
preprocessing removes uncharacterized entries and retains one lowest-energy
conformation per molecule, leaving 130,831 molecules split into 100,000 for
training, 17,748 for validation, and 13,083 for testing. Molecules contain at
most 29 atoms when hydrogens are included.

\paragraph{GEOM-DRUG.}
GEOM-DRUG~\citep{axelrod2022geom} contains larger and more flexible drug-like
molecules. Starting from approximately 304K molecules, the EDM preprocessing
retains up to the 30 lowest-energy conformations per molecule, yielding roughly
six million conformer records. Our processed data contain 5,538,014 training
and 692,251 validation records. Molecules have 44.2 atoms on average and up to
181 atoms. Both datasets include explicit hydrogen atoms.

\section{Implementation Details}
\label{app:sec:implementation}

\subsection{Data Preprocessing}
\label{app:data_preprocessing}

We embed each molecule in a cubic domain of side length $B$, using
$B=15.36$\,\AA{} for QM9 and $B=30.72$\,\AA{} for GEOM-DRUG.
The atom-type vocabularies are \{H, C, N, O, F\} for QM9 and
\{H, B, C, N, O, F, Al, Si, P, S, Cl, As, Br, I, Hg, Bi\} for GEOM-DRUG.
For each conformation, we translate the atomic coordinates by their centroid
and place the centered molecule in $[-B/2,B/2)^3$. If it exceeds the domain,
we recenter it on the atom nearest the centroid and retain only atoms inside
the box. During training, a uniformly sampled random 3D rotation is applied
before voxelization for the joint atom flow only; the hierarchical occupancy
flow is trained without rotation augmentation.

Our main experiments use $D=6$ and $k=2$, corresponding to occupancy
generation at depths 2, 4, and 6 and a final $64^3$ grid. The resulting cell
widths are $0.24$\,\AA{} for QM9 and $0.48$\,\AA{} for GEOM-DRUG.
The effects of the final depth $D$ and depth increment $k$ are studied in
\cref{sec:hierarchy_ablation}.

\subsection{Cell Corruption}
\label{app:cell_corruption}

We select up to $\lfloor N/10\rfloor$ atoms uniformly without replacement.
Each selected cell is assigned a shift
$\mathbf{s}\in\{\pm\mathbf{e}_1,\pm\mathbf{e}_2,\pm\mathbf{e}_3\}$,
where $\mathbf{e}_i$ are the Cartesian unit vectors. Shifts outside the domain
or into occupied cells are skipped. For cell center $\mathbf{c}_j$ and local
displacement $\boldsymbol{\Delta}_j=r_j\mathbf{q}_j$, the update
$(\mathbf{c}_j,\boldsymbol{\Delta}_j)\mapsto
(\mathbf{c}_j+h_D\mathbf{s},\boldsymbol{\Delta}_j-\mathbf{s})$
preserves the coordinate in \cref{eq:molecule_feature}. We re-encode the
compensated displacement as length and direction. The number of proposed
shifts is sampled uniformly from zero to $\lfloor N/10\rfloor$.

\subsection{Neural Network Architecture}
\label{app:network_architecture}

\begin{figure}[t]
  \centering
  \includegraphics[width=0.7\linewidth]{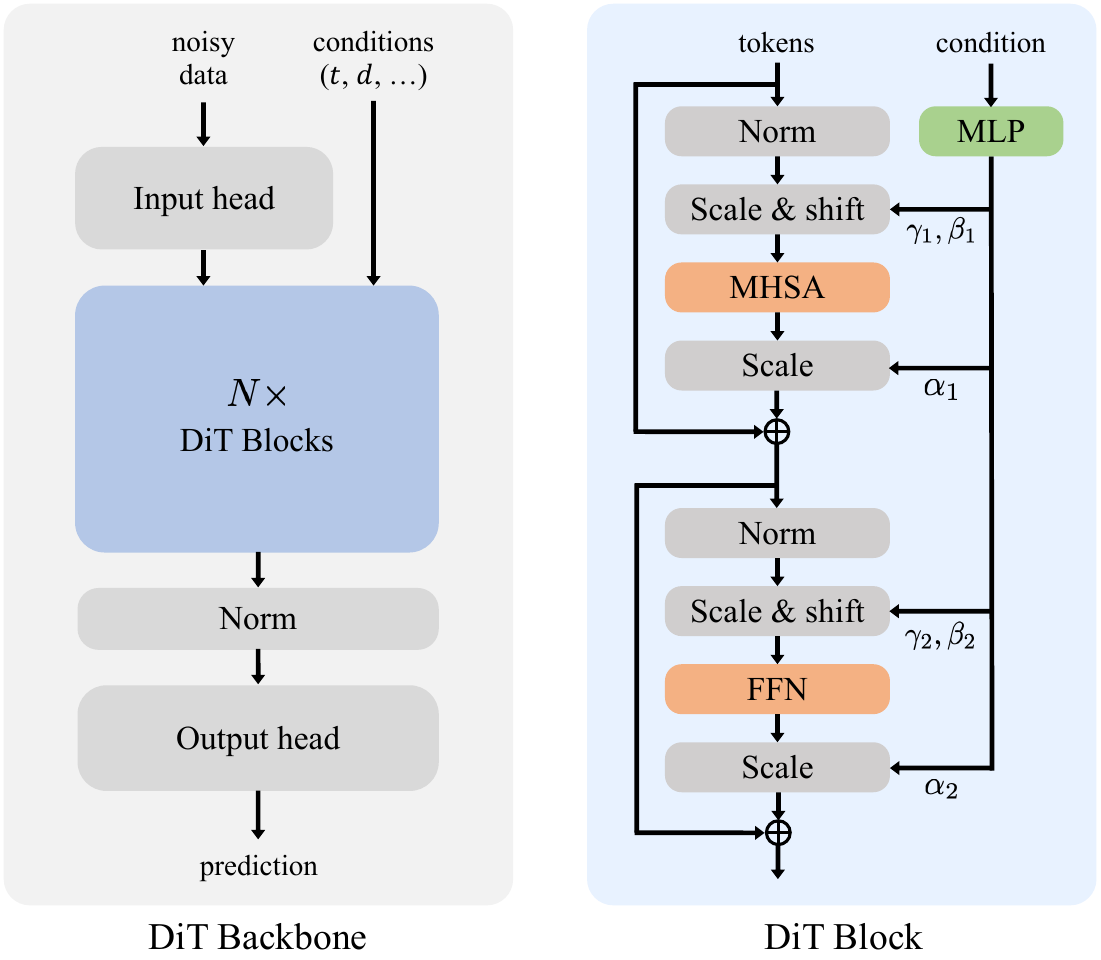}
  \caption{Sparse transformer backbone used in OctMole. Cell coordinates
  provide spatial positional encoding, while time and, where applicable,
  target-depth conditions modulate the transformer blocks. Task-specific input
  and output heads instantiate the hierarchical occupancy flow or joint atom
  flow. Coordinate rotary embeddings and the register tokens are omitted from
  the schematic for clarity.}
  \label{fig:backbone}
\end{figure}

\paragraph{Shared backbone.}
Both flows use the sparse DiT-style backbone illustrated in
\cref{fig:backbone}. Each block consists of non-causal multi-head
self-attention and a SwiGLU feed-forward network, with RMS normalization and
adaptive shift, scale, and residual gates produced from the conditioning
embedding. Cell indices enter attention through three-dimensional rotary
position embeddings, while flow time modulates every block; the occupancy
flow additionally adds an embedding of the target octree depth. We insert
eight learned register tokens partway through the transformer and remove them
before the output heads. The final task heads are two-layer MLPs with a GELU
activation and a zero-initialized output projection.

\paragraph{Flow-specific inputs and outputs.}
For the hierarchical occupancy flow with $k=2$, each sparse token represents
an occupied parent cell. Its input contains the noisy occupancy of the 64
candidate descendants together with the previous clean prediction used for
self-conditioning, and its output head predicts the 64 clean endpoint values.
For the joint atom flow, the input is the sum of an atom-type embedding and a
linear embedding of the four continuous local features $(r,\mathbf{q})$.
Before cell corruption, the radial offset is normalized by the finest-cell
width:
\[
  r_j=\frac{\lVert\mathbf{x}_j-\mathbf{c}_j\rVert_2}{h_D},
  \qquad r_j\in[0,\sqrt{3}/2].
\]
Here, $\mathbf{x}_j$ is the atom position, $\mathbf{c}_j$ is the center of its
finest occupied cell, and $h_D$ is the cell width.
Separate heads predict endpoint atom-type probabilities and continuous local
features from the same molecular context. At inference, the four predicted
continuous features are decoded directly with \cref{eq:molecule_feature},
without clipping the length or renormalizing the direction.

\paragraph{Model configurations.}
The width, attention-head dimension, and MLP expansion ratio are 768, 64, and
4 for all models, with no attention or projection dropout. The occupancy flows
and the QM9 atom flow use 12 blocks and insert their registers after the fourth
block; the GEOM-DRUG atom flow uses 16 blocks and inserts them after the fifth.
Dataset-specific configurations are summarized in
\cref{tab:architecture_details}.

\begin{table}[H]
  \caption{Backbone configurations used in the main experiments. Parameter
  counts include the task-specific input and output heads.}
  \label{tab:architecture_details}
  \tablestyle{5pt}{1.1}
  \begin{tabular}{llcccc}
    \toprule
    \textbf{Dataset} & \textbf{Flow} & \textbf{Blocks} & \textbf{Heads} &
    \textbf{Registers} & \textbf{Parameters} \\
    \midrule
    QM9 & Occupancy & 12 & 12 & 8 & 129.11M \\
    QM9 & Atom & 12 & 12 & 8 & 129.74M \\
    GEOM-DRUG & Occupancy & 12 & 12 & 8 & 129.11M \\
    GEOM-DRUG & Atom & 16 & 12 & 8 & 172.26M \\
    \bottomrule
  \end{tabular}
\end{table}

\subsection{Training}

We train with AdamW using $\beta_1=0.9$, $\beta_2=0.95$, zero weight decay,
and bfloat16 mixed precision. The learning rate is linearly warmed up for five
epochs to $2\times10^{-4}$ and then held constant. Flow times are obtained by
applying a sigmoid to samples from $\mathcal{N}(-0.8,0.8^2)$; Gaussian
coordinate and occupancy noise have unit scale, and we use
$\varepsilon=0.05$. The occupancy flow uses self-conditioning with probability
$0.5$: the first endpoint prediction is detached and supplied to a second
network evaluation; otherwise a zero tensor is used. For the atom flow, the
continuous-feature and atom-type loss weights
are $1$ and $0.1$, respectively. We apply random rotations when training the
atom flow on both datasets, and apply cell corruption only on GEOM-DRUG for
the main results. The occupancy-flow target depth is sampled uniformly
from depths 2, 4, and 6. All models use an effective batch size of 1,024, and the
training configurations used for sampling are listed in
\cref{tab:training_details}.

\begin{table}[H]
  \caption{Training configurations used for the main results. Epoch is the
  training epoch used at inference. All GPUs are NVIDIA
  A100-SXM4-40GB.}
  \label{tab:training_details}
  \tablestyle{5pt}{1.1}
  \begin{tabular}{llcccc}
    \toprule
    \textbf{Dataset} & \textbf{Flow} & \textbf{GPUs} &
    \textbf{Batch/GPU} & \textbf{Epoch} & \textbf{Cell corruption} \\
    \midrule
    QM9 & Occupancy & 4 & 256 & 4000 & -- \\
    QM9 & Atom & 4 & 256 & 1000 & No \\
    GEOM-DRUG & Occupancy & 8 & 128 & 100 & -- \\
    GEOM-DRUG & Atom & 8 & 128 & 10 & Yes \\
    \bottomrule
  \end{tabular}
\end{table}

\subsection{Sampling}

We sample the exponential moving average of each model with decay $0.9999$.
The occupancy flow generates depths 2, 4, and 6 using 50 uniformly spaced
steps with Heun integration and a final Euler update, thresholding the endpoint
at zero after each depth. During inference, self-conditioning is initialized to
zero and replaced after each integration step by the detached endpoint
prediction from the previous step. Conditioned on the generated depth-6 support, the
atom flow uses 50 updates that advance continuous local features by Euler
integration. For a current type $a_i$ at times $t_i<t_{i+1}$, let
$\pi_i(b)$ be the predicted endpoint probability of type $b$. The categorical
update assigns probability
$\frac{t_{i+1}-t_i}{1-t_i}\pi_i(b)$ to each $b\ne a_i$ and assigns the
remaining probability to staying at $a_i$. We
sample with seed 0 on four GPUs, using 256 molecules per GPU for QM9 and 128
for GEOM-DRUG. Candidates are ranked by the mean log probability of their
sampled atom types,
\[
  s(M)=\frac{1}{N}\sum_{j=1}^{N}\log p_\phi(a_j\mid
  A_t,U_t,\mathcal{S}_D,t),
\]
evaluated from the final atom-type prediction. Within each consecutive group
of 512 candidates, retaining the top 10\% for QM9 and 20\% for GEOM-DRUG
requires 98,464 and 49,709 candidates, respectively, to obtain the 10,000
final molecules reported in the main experiments. This chunked confidence
selection follows the same basic protocol as Uni-3DAR~\citep{Uni3DAR2025}.

For spatial conditioning at depth $d_m$, we project the reference leaves to
their unique depth-$d_m$ ancestors, use these cells in place of the generated
state at that depth, and run only the remaining occupancy stages. At
$d_m=D$, the occupancy flow is bypassed and the supplied leaves are passed
directly to the atom flow. For regional editing, retained depth-$D$ leaves are
united with leaves regenerated from the editable coarse cells, with duplicate
cells removed, before the atom flow regenerates atom types and local offsets
over the combined support.

\section{Evaluation Protocols}
\label{app:sec:evaluation}

\subsection{Open Babel Post-processing}
Our model generates atom types and Cartesian coordinates without an explicit
molecular graph. Following VoxMol~\citep{pinheiro2023voxmol}, we first remove
atoms involved in unrealistically close contacts (below $0.8$\,\AA). We then
write the remaining typed point cloud as an XYZ structure, use
Open Babel 3.2.1~\citep{oboyle2011openbabel} to infer connectivity and bond orders, and
perform an XYZ-to-SDF round trip. The resulting SDF record is converted into
an RDKit molecule for all subsequent chemical checks. We apply exactly the
same conversion to the validation set when constructing reference
distributions.

Stability is evaluated on the inferred graph before sanitization. For validity,
we separate disconnected fragments, retain the largest fragment, sanitize it
with RDKit, and obtain its canonical SMILES representation. Connectedness is
recorded before fragment selection. Failed conversions or sanitization are
counted as invalid.

\subsection{Chemical Quality Metrics}
Let $\mathcal{G}$ be the set of converted molecules, $A(G)$ the atoms of a
molecule, and $s(a)$ an indicator that the inferred valency of atom $a$ belongs
to the allowed valencies for its element. Atom and molecule stability are
defined as
\begin{equation}
  \mathrm{AtomSta.} =
  \frac{\sum_{G\in\mathcal{G}}\sum_{a\in A(G)}s(a)}
       {\sum_{G\in\mathcal{G}}|A(G)|}, \qquad
  \mathrm{MolSta.} =
  \frac{1}{|\mathcal{G}|}\sum_{G\in\mathcal{G}}
  \prod_{a\in A(G)}s(a).
\end{equation}
Validity is the fraction of converted samples whose largest fragment can be
sanitized by RDKit. Uniqueness is the fraction of distinct canonical SMILES
among valid samples, while connectedness is the fraction of converted samples
that contain exactly one connected component.

\subsection{Distribution Metrics}
We construct empirical distributions over the number of atoms, atom types,
bond types, element-conditioned valencies, bond-order-conditioned bond
lengths, and center-element-conditioned bond angles. Categorical atom- and
bond-type distributions are compared using total variation distance,
\begin{equation}
  \operatorname{TV}(p,q)=\frac{1}{2}\sum_i |p_i-q_i|.
\end{equation}
The remaining one-dimensional distributions are compared using the
Wasserstein-1 distance,
\begin{equation}
  W_1(p,q)=\int_{-\infty}^{\infty}|F_p(z)-F_q(z)|\,\mathrm{d}z,
\end{equation}
where $F_p$ and $F_q$ denote empirical cumulative distribution functions. We
use integer bins for molecule size, $0.01$\,\AA{} bins for bond lengths, and
$0.1^\circ$ bins for bond angles. Conditional distances are averaged using
their frequencies in the reference set. The resulting metrics are reported as
Nodes W1, Atom TV, Bond TV, Valency W1, Length W1, and Angle W1.

\section{Main Experiments}
\label{app:sec:experiments}
\subsection{Distributional Fidelity}
\label{app:distribution_fidelity}

\Cref{app:tab:openbabel_distribution} reports all six distributional
distances for the 10,000-sample sets used in
\cref{tab:openbabel_quality}. On GEOM-DRUG, OctMole is closer to the reference
than Uni-3DAR in molecular size, atom type, bond type, valency, and bond angle;
Uni-3DAR has the lower bond-length distance. On QM9, OctMole is closer than
Uni-3DAR in all six metrics, while its atom-type distance remains higher than
EDM, GeoLDM, and UniGEM.

\begin{table*}[t]
  \caption{Distributional fidelity under our Open Babel evaluation. Lower is
  better for every metric. TV denotes total variation distance and $W_1$
  denotes one-dimensional Wasserstein distance. \ranknote}
  \label{app:tab:openbabel_distribution}
  \tablestyle{2.5pt}{1.12}
  \begin{tabular}{lcccccc}
    \toprule
    Method & Nodes $W_1$ & Atom TV & Bond TV & Valency $W_1$ & Length $W_1$ & Angle $W_1$ \\
    \midrule
    \multicolumn{7}{c}{\textbf{QM9}} \\
    \midrule
    EDM~\citep{hoogeboom2022equivariant} & -- & .013020 & \secondcell{.000711} & .011710 & \bestcell{.000548} & .441630 \\
    GeoLDM~\citep{xu2023geometric} & -- & \bestcell{.009140} & \bestcell{.000689} & .004940 & \secondcell{.000716} & \bestcell{.398670} \\
    UniGEM$^{*}$~\citep{feng2024unigem} & -- & \secondcell{.009920} & .005794 & .008080 & .002367 & 1.472070 \\
    Uni-3DAR~\citep{Uni3DAR2025} & \secondcell{.859498} & .027621 & .006703 & \secondcell{.001959} & .001845 & .579394 \\
    OctMole & \bestcell{.316405} & .022172 & .002656 & \bestcell{.000929} & .000805 & \secondcell{.437003} \\
    \midrule
    \multicolumn{7}{c}{\textbf{GEOM-DRUG}} \\
    \midrule
    EDM~\citep{hoogeboom2022equivariant} & -- & .045000 & .002550 & .139500 & \secondcell{.001500} & 5.547000 \\
    GeoLDM~\citep{xu2023geometric} & -- & .050800 & \bestcell{.001710} & .085300 & \bestcell{.001310} & 2.991000 \\
    UniGEM$^{*}$~\citep{feng2024unigem} & -- & .055300 & \secondcell{.001890} & .092600 & .001850 & 3.934000 \\
    Uni-3DAR~\citep{Uni3DAR2025} & \secondcell{16.580815} & \secondcell{.041273} & .022111 & \secondcell{.022945} & .001712 & \secondcell{1.026799} \\
    OctMole & \bestcell{2.984420} & \bestcell{.014374} & .003236 & \bestcell{.008375} & .001962 & \bestcell{.891154} \\
    \bottomrule
  \end{tabular}
\end{table*}

\subsection{GFN2-xTB Evaluation}
\label{app:xtb_evaluation}

We use GFN2-xTB as implemented in xTB 6.7.1. Formal charges and
unpaired-electron counts are inferred with RDKit from SMILES derived with
Open Babel. After the same clash-atom removal as in the chemical evaluation, we
perform a single-point calculation followed by geometry optimization at xTB's
\texttt{normal} convergence level, with time limits of 60 and 180 seconds,
respectively. An optimization is converged only if xTB exits successfully,
reports normal termination and geometry convergence, and returns a final
geometry and energy. RMSD uses fixed atom correspondence after centering and
Kabsch alignment with a proper rotation; reflections, atom permutations, and
symmetry matching are not considered.

Relaxation energy per atom is
$(E_{\mathrm{initial}}-E_{\mathrm{optimized}})/N$ in kcal/mol/atom.
GEOM-DRUG reference conformers are already near GFN2-xTB minima, whereas QM9
reference geometries come from a different electronic-structure level.
\Cref{app:tab:xtb_full} reports convergence among valid, connected inputs,
end-to-end yield among all 10,000 final samples, and median and 95th-percentile
energy and RMSD among converged molecules.

\begin{table*}[t]
  \caption{Full GFN2-xTB comparison from 10,000 final samples per model.
  $N_{\rm vc}$ counts valid, connected inputs; convergence is relative to
  these inputs and yield to all final samples. Energy is in kcal/mol/atom and
  RMSD in \AA; both report median (95th percentile).}
  \label{app:tab:xtb_full}
  \tablestyle{3pt}{1.12}
  \begin{tabular}{lrrrrr}
    \toprule
    Method & $N_{\rm vc}$ & Conv. (\%) & Yield (\%) & Energy & RMSD \\
    \midrule
    \multicolumn{6}{c}{\textbf{QM9}} \\
    \midrule
    Data & 10,000 & 100.00 & -- & 0.0690 (0.1431) & 0.0574 (0.2589) \\
    EDM & 9,839 & 100.00 & 98.39 & 0.1613 (1.1166) & 0.2117 (0.8272) \\
    GeoLDM & 9,880 & 100.00 & 98.80 & 0.1001 (0.5513) & 0.1174 (0.6952) \\
    UniGEM & 9,888 & 100.00 & 98.88 & 0.1194 (0.8162) & 0.1364 (0.7021) \\
    Uni-3DAR & 9,948 & 99.99 & 99.47 & 0.2178 (0.6282) & 0.2349 (0.8173) \\
    OctMole & 9,954 & 99.99 & 99.53 & 0.1149 (0.5162) & 0.0628 (0.2568) \\
    \midrule
    \multicolumn{6}{c}{\textbf{GEOM-DRUG}} \\
    \midrule
    Data & 10,000 & 100.00 & -- & $\approx0$ ($\approx0$) & $\approx0$ ($\approx0$) \\
    EDM & 3,628 & 96.86 & 35.14 & 3.6000 (7.9797) & 1.3819 (2.2637) \\
    GeoLDM & 4,570 & 98.16 & 44.86 & 1.7871 (5.0949) & 1.2544 (2.1846) \\
    UniGEM & 3,724 & 98.25 & 36.59 & 1.7689 (5.2115) & 1.2475 (2.1557) \\
    Uni-3DAR & 8,667 & 99.82 & 86.51 & 0.2988 (1.8673) & 0.6019 (1.4218) \\
    OctMole & 9,514 & 99.71 & 94.86 & 0.6335 (1.2050) & 0.6067 (1.4693) \\
    \bottomrule
  \end{tabular}
\end{table*}

\subsection{Molecular Size and Physical Diagnostics}
\label{app:size_diagnostics}

\begin{figure}[t]
  \centering
  \includegraphics[width=0.75\linewidth]{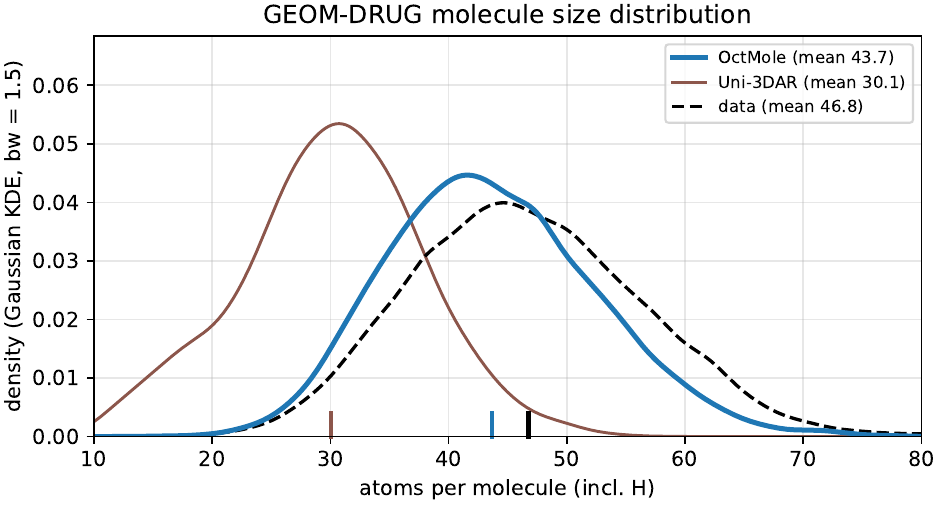}
  \caption{GEOM-DRUG atom-count distributions, including hydrogens, for
  10,000 raw samples from each generator and a uniform 10,000-molecule sample
  from the training split. Curves are Gaussian kernel-density estimates;
  Uni-3DAR is shifted toward smaller molecules, whereas OctMole tracks the
  data distribution.}
  \label{app:fig:size_distribution}
\end{figure}

Uni-3DAR preferentially generates smaller GEOM-DRUG molecules: mean atom
counts, including hydrogens, are 30.1 for Uni-3DAR, 43.7 for OctMole, and 46.8
for the data (\cref{app:fig:size_distribution}), which may explain its low
aggregate relaxation energy.

\begin{figure}[t]
  \centering
  \includegraphics[width=0.85\linewidth,trim=0 0 0 20bp,clip]{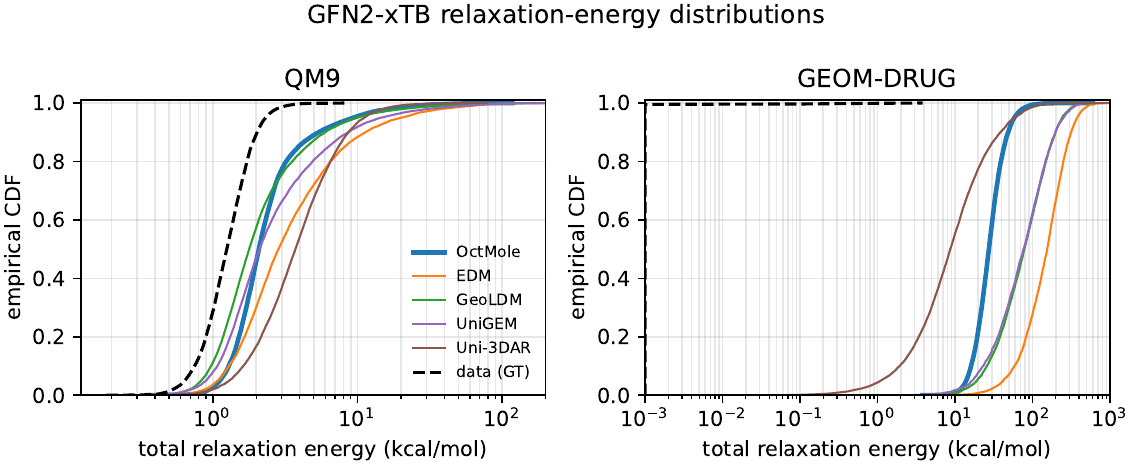}
  \caption{Empirical cumulative distributions of total GFN2-xTB relaxation
  energy over the full 10,000-sample evaluations, restricted to valid,
  connected molecules with converged optimization. The horizontal axis is
  logarithmic; nonpositive reference values are displayed at
  $10^{-3}$\,kcal/mol.}
  \label{app:fig:strain_distribution}
\end{figure}

The empirical relaxation-energy distributions in
\cref{app:fig:strain_distribution} diagnose geometric strain rather than
equilibrium populations across molecules. A Boltzmann factor
$\exp[-\Delta E/(k_{\rm B}T)]$ illustrates why large relaxation energies are
unfavorable, but molecules of different composition cannot be compared as a
shared thermodynamic ensemble.

\begin{figure}[t]
  \centering
  \includegraphics[width=\linewidth]{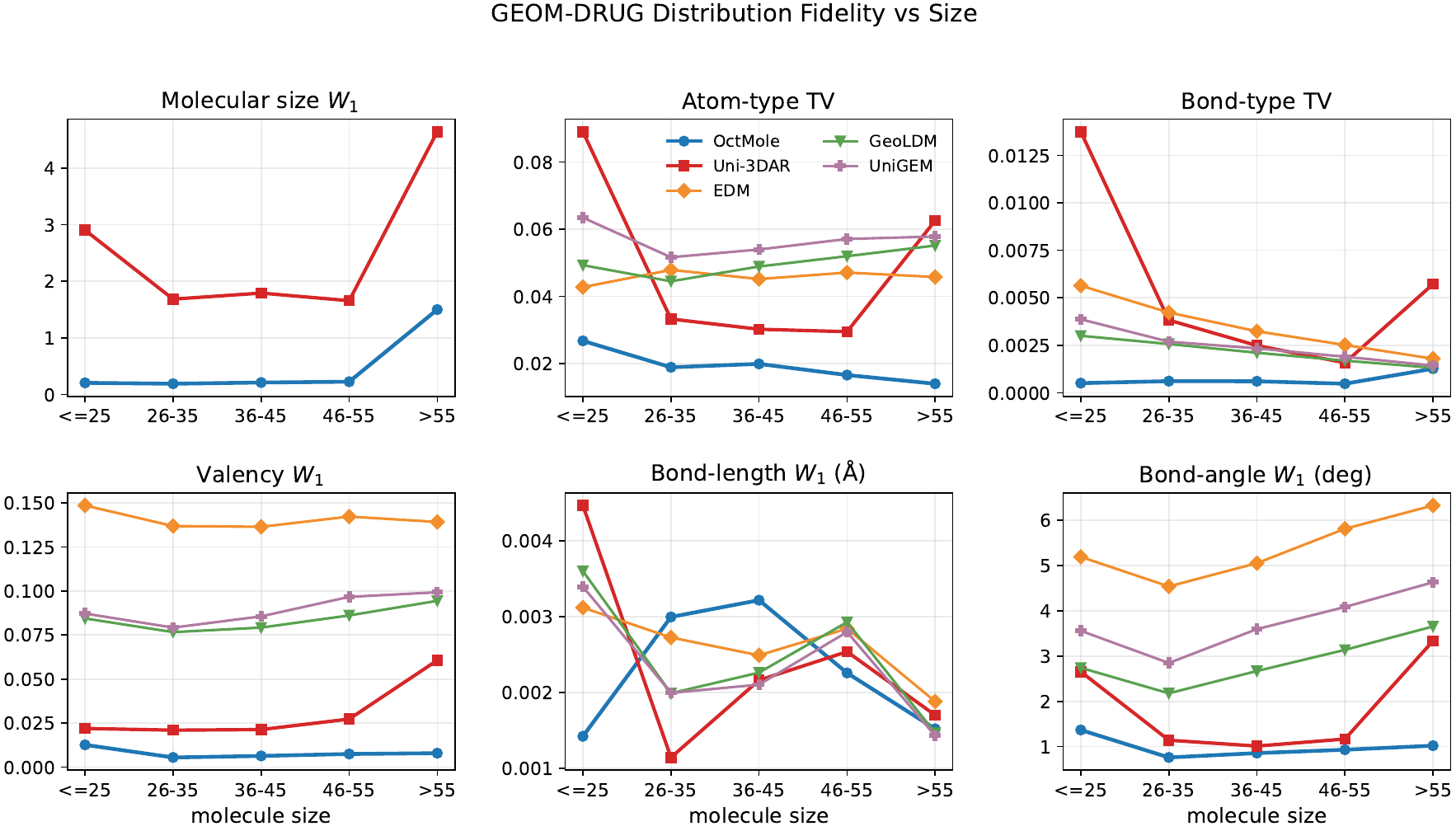}
  \caption{GEOM-DRUG distributional distances stratified by molecular size.
  Across all five size bins, OctMole is closest to the reference in molecular
  size, atom type, bond type, valency, and bond angle; bond length is the sole
  exception. Lower values indicate closer agreement with the reference.}
  \label{app:fig:distribution_vs_size}
\end{figure}

As shown in the main-text \cref{fig:stability_vs_size}, OctMole has the highest
GEOM-DRUG molecule stability in every size bin among the evaluated generators.
Its stability stays between 85.7\% and 93.6\%, while Uni-3DAR declines from
85.3\% for molecules with at most 25 atoms to 62.8\% at 46--55 atoms.
\Cref{app:fig:distribution_vs_size} shows that this advantage extends beyond
stability: in every size bin, OctMole has the lowest distance among the
evaluated generators for molecular size, atom type, bond type, valency, and
bond angle. Bond length is the only metric on which it is not consistently
best.

\subsection{Complete Multiscale Conditioning Results}
\label{app:multiscale_control}

The unified conditioning protocol uses the same 256 references per dataset,
candidate budgets of 9,943 on GEOM-DRUG and 19,696 on QM9, and confidence
reranking to 2,000 final samples per depth. IoU is measured against the
reference occupancy, Chamfer distance includes hydrogens without rotational
alignment, and chemical similarity uses radius-2, 2,048-bit Morgan fingerprints
of SMILES inferred with Open Babel.

\begin{table*}[t]
  \caption{Complete multiscale conditioning results. Joint is the percentage
  of finals that are valid and stable, within 1\,\AA{} Chamfer of the
  reference, and chemically distinct from it, with a 95\% cluster-bootstrap
  interval over references.}
  \label{app:tab:multiscale_control_full}
  \tablestyle{5pt}{1.12}
  \begin{tabular}{llrrrrrrl}
    \toprule
    Dataset & Depth & IoU@$d$ & IoU@6 & Chamfer (\AA) & Sim. & Div. & Sta. & Joint \\
    \midrule
    QM9 & 2 & 1.00 & 0.00 & 1.00 & 0.12 & 0.88 & 99.7 & 41.3 [33.6, 49.1] \\
        & 4 & 0.97 & 0.51 & 0.08 & 0.97 & 0.01 & 100.0 & 4.2 [0.4, 10.4] \\
        & 6 & 0.99 & 0.99 & 0.03 & 1.00 & 0.00 & 100.0 & 0.0 [0.0, 0.0] \\
    \midrule
    GEOM-DRUG & 2 & 0.99 & 0.01 & 1.33 & 0.13 & 0.88 & 85.2 & 1.1 [0.5, 1.9] \\
              & 4 & 0.98 & 0.34 & 0.27 & 0.74 & 0.28 & 95.4 & 53.3 [45.5, 61.5] \\
              & 6 & 1.00 & 1.00 & 0.06 & 1.00 & 0.00 & 99.1 & 1.7 [0.5, 3.5] \\
    \bottomrule
  \end{tabular}
\end{table*}

\begin{figure}[t]
  \centering
  \includegraphics[width=0.55\linewidth]{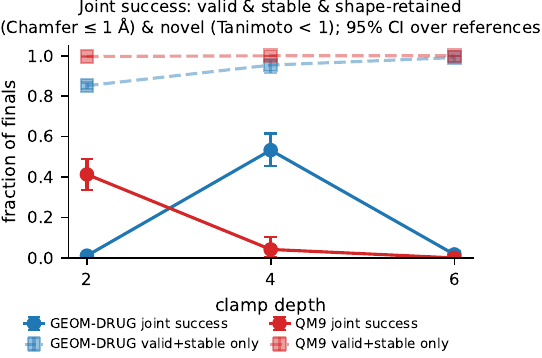}
  \caption{Joint success by conditioning depth, with 95\% cluster-bootstrap
  intervals. The useful depth shifts from 2 on QM9 to 4 on GEOM-DRUG.}
  \label{fig:multiscale_joint_success}
\end{figure}

\Cref{app:tab:multiscale_control_full,fig:multiscale_joint_success} show that
the most useful conditioning depth depends on molecular scale. Depth-6
reranking concentrates the final samples on 88 of the 256 GEOM-DRUG
references, with as many as 40 finals from one reference. The clustered
bootstrap intervals account for covered-reference concentration but cannot
recover references absent after selection.

\subsection{Rotation and Grid-Origin Sensitivity}
\label{app:coordinate_sensitivity}

We randomly sample 256 validation molecules per dataset and apply either no
transformation, a random SO(3) rotation, a random grid-origin shift within half
a finest-cell width per axis, or both rotation and shift. We then re-voxelize
each molecule, truncate its octree at depth 2, 4, or 6, and use that occupancy
to condition generation; outputs are inverse-transformed before evaluation.
Each setting retains 2,000 samples. The same references and four random draws
per reference are shared across depths and model variants.
\Cref{app:tab:rotation} reports the resulting molecule stability.

\begin{table}[t]
  \centering
  \caption{Molecule stability (\%) under coordinate transformations with cell
  corruption enabled. Shift changes the grid origin; Both combines shift and
  rotation.}
  \label{app:tab:rotation}
  \tablestyle{6pt}{1.12}
  \begin{tabular}{lrrrrr}
    \toprule
    Dataset & Depth & Identity & Rotation & Shift & Both \\
    \midrule
    QM9 & 2 & 99.65 & 99.80 & 99.70 & 99.70 \\
        & 4 & 100.00 & 99.50 & 100.00 & 98.65 \\
        & 6 & 100.00 & 100.00 & 100.00 & 99.95 \\
    \midrule
    GEOM-DRUG & 2 & 85.20 & 86.90 & 85.40 & 85.00 \\
              & 4 & 95.40 & 82.95 & 93.05 & 80.05 \\
              & 6 & 99.10 & 97.05 & 98.15 & 94.65 \\
    \bottomrule
  \end{tabular}
\end{table}

For GEOM-DRUG depth 4, connectedness also falls from 99.70\% to 93.15\%
under rotation and to 92.10\% under both transformations. Depths 2 and 6 are
less affected, and supplied-cell coverage remains above 99.99\%. On QM9,
cell corruption limits the depth-4 combined-transformation stability drop to
1.35 points, compared with 21.35 points without corruption. The experiment
measures re-quantization and generation together rather than establishing
equivariance.

\section{Ablation Studies}
\label{app:sec:ablations}
\subsection{Sparse and Dense System Details}
\label{app:sparse_details}

The dense and octree systems use the same final $64^3$ resolution, number of
epochs, GPUs, and 50-step Heun sampling within each dataset. Their spatial
generators, parameter counts, effective batch sizes, and architectures differ;
the comparison therefore measures complete systems rather than sparsity in
isolation.
\Cref{app:tab:dense_training,app:tab:dense_sampling} report the corresponding
training and sampling measurements.

\begin{table*}[t]
  \caption{Training efficiency of dense and octree spatial generators. Batch
  is per GPU and time is observed wall-clock training time.}
  \label{app:tab:dense_training}
  \tablestyle{5pt}{1.12}
  \begin{tabular}{llrrrrrr}
    \toprule
    Dataset & Representation & Params (M) & GPUs & Batch & Epochs & Time (h) & Speedup \\
    \midrule
    QM9 & Dense & 131.31 & 4 & 32 & 4,000 & 118.84 & $1.00\times$ \\
        & Octree & 129.10 & 4 & 256 & 4,000 & 17.35 & $6.85\times$ \\
    \midrule
    GEOM-DRUG & Dense & 236.07 & 8 & 32 & 100 & 143.86 & $1.00\times$ \\
              & Octree & 129.10 & 8 & 128 & 100 & 28.02 & $5.13\times$ \\
    \bottomrule
  \end{tabular}
\end{table*}

For inference, each method retains its complete candidate-generation and
reranking procedure to produce approximately 1,000 final molecules on one GPU.
Timing excludes subsequent Open Babel evaluation.

\begin{table*}[t]
  \caption{Single-GPU sampling efficiency and quality for approximately 1,000
  reranked final samples. Time is seconds per retained molecule. Chemical
  metrics are percentages; lower Nodes $W_1$ is better.}
  \label{app:tab:dense_sampling}
  \tablestyle{4pt}{1.12}
  \begin{tabular}{llrrrrrrr}
    \toprule
    Dataset & Repr. & Time & Atom Sta. & Mol. Sta. & Valid & Unique & Conn. & Nodes $W_1$ \\
    \midrule
    QM9 & Dense & 2.090 & 96.831 & 69.500 & 94.300 & 99.576 & 80.400 & 3.5609 \\
        & Octree & 0.379 & 99.989 & 99.900 & 99.900 & 98.298 & 100.000 & 0.5004 \\
    \midrule
    GEOM-DRUG & Dense & 1.928 & 49.197 & 4.911 & 99.896 & 22.176 & 34.169 & 42.4219 \\
              & Octree & 0.443 & 99.543 & 84.400 & 98.200 & 100.000 & 96.800 & 2.6977 \\
    \bottomrule
  \end{tabular}
\end{table*}

\begin{figure}[t]
  \centering
  \includegraphics[width=0.9\linewidth]{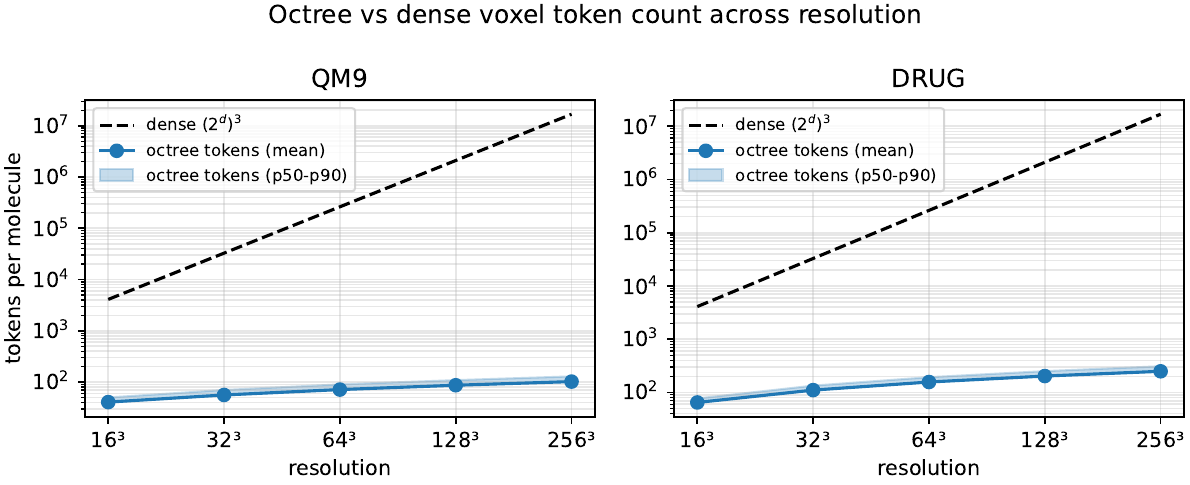}
  \caption{Tokens per molecule across equivalent resolutions for 1,000
  validation molecules from each dataset. Octree counts sum occupied nodes
  over all levels; lines show means and shaded bands show the 50th--90th
  percentiles. Dense-grid token counts grow cubically, whereas sparse octree token counts
  grow approximately linearly.}
  \label{app:fig:resolution_cost}
\end{figure}

From $64^3$ to $256^3$, dense grids grow by $64\times$, whereas mean octree
node counts grow by only $1.59\times$ on GEOM-DRUG and $1.43\times$ on QM9.
Measured occupancy-flow sampling from depth 6 to 8 on GEOM-DRUG grows from
0.079 to 0.110 seconds per molecule and from 691 to 776 MB peak memory. The
number of generation stages can dominate token count
(\cref{app:fig:resolution_cost}): QM9 depth 5 is slower
than depth 6 because its $k=1$ schedule has five stages rather than three.

\subsection{Continuous-Offset Bond Geometry}
\label{app:offset_geometry}

\begin{figure}[t]
  \centering
  \includegraphics[width=0.8\linewidth]{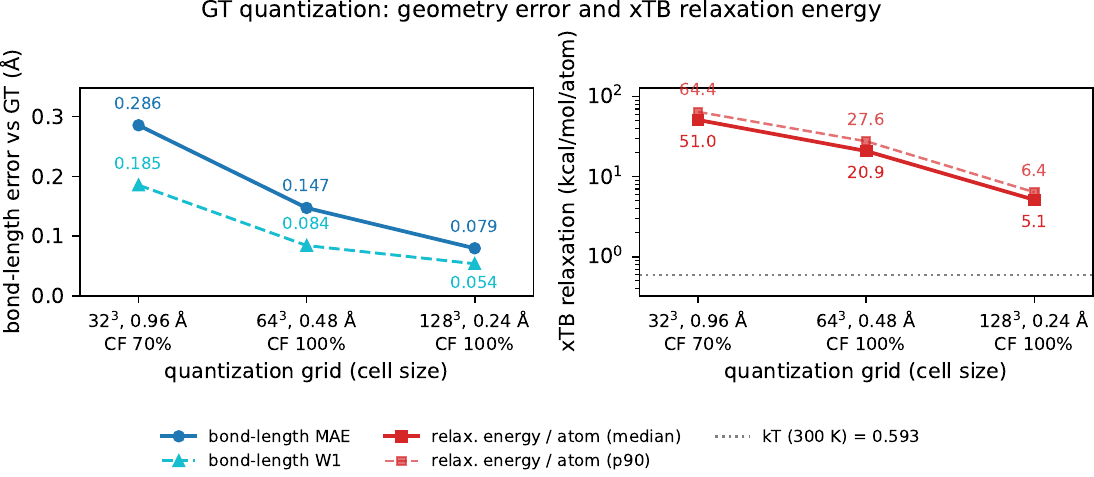}
  \caption{Quantization strain for 300 ground-truth GEOM-DRUG validation
  conformers at $32^3$, $64^3$, and $128^3$ resolution without continuous
  offsets. Collision-bearing conformers are counted in the collision-free
  rate and excluded from geometry and xTB summaries at that resolution. The
  working $64^3$ grid has no collisions in this cohort but still incurs
  substantial bond error and GFN2-xTB relaxation energy.}
  \label{fig:quantization_xtb}
\end{figure}

We snap generated coordinates to their assigned cell centers while keeping
support and atom types fixed. This inference intervention isolates the
geometric contribution of continuous offsets, rather than training a separate
model. Distances are computed per bond or angle type and aggregated with
reference type frequencies. Snapping produces grid-induced errors and raises
GEOM-DRUG bond-length $W_1$ from 0.017 to 0.135\,\AA{} and bond-angle $W_1$
from 1.12 to 7.97 degrees. QM9 bond-length $W_1$ rises from 0.004 to
0.068\,\AA{}. Direct quantization of validation conformers provides a
model-independent view of the same effect in \cref{fig:quantization_xtb}.

\subsection{Hierarchy Settings}
\label{app:hierarchy_settings}

We compare $k=1$ and $k=2$ at fixed finest depth $D=6$, using six and three
occupancy stages, respectively. We additionally evaluate $D=8,k=2$ on
GEOM-DRUG. The atom model and 50-step sampling method are fixed within each
dataset.

At $D=6$, $k=1$ and $k=2$ have comparable chemical quality, while $k=2$
halves the number of stages. Increasing to $D=8$ without retuning substantially
degrades the GEOM-DRUG result (\cref{app:tab:hierarchy_ablation}).

\begin{table*}[t]
  \caption{Hierarchy ablation on 10,000 reranked final samples. Chemical
  metrics are percentages; lower $W_1$ is better.}
  \label{app:tab:hierarchy_ablation}
  \tablestyle{3pt}{1.12}
  \begin{tabular}{lccrrrrrrr}
    \toprule
    Dataset & $D$ & $k$ & Atom Sta. & Mol. Sta. & Valid & Connected & Nodes $W_1$ & Length $W_1$ \\
    \midrule
    QM9 & 6 & 1 & 99.935 & 99.080 & 99.560 & 99.700 & 0.9984 & 0.0015 \\
        & 6 & 2 & 99.985 & 99.730 & 99.770 & 99.960 & 0.3183 & 0.0008 \\
    \midrule
    GEOM-DRUG & 6 & 1 & 99.614 & 86.790 & 98.180 & 96.960 & 3.6906 & 0.0023 \\
              & 6 & 2 & 99.624 & 86.820 & 98.200 & 96.900 & 2.9844 & 0.0020 \\
              & 8 & 2 & 98.229 & 52.700 & 92.440 & 82.160 & 4.9498 & 0.0039 \\
    \bottomrule
  \end{tabular}
\end{table*}

\subsection{Local Perturbations across Depths}
\label{app:local_perturbations}

\begin{figure}[t]
  \centering
  \includegraphics[width=\linewidth]{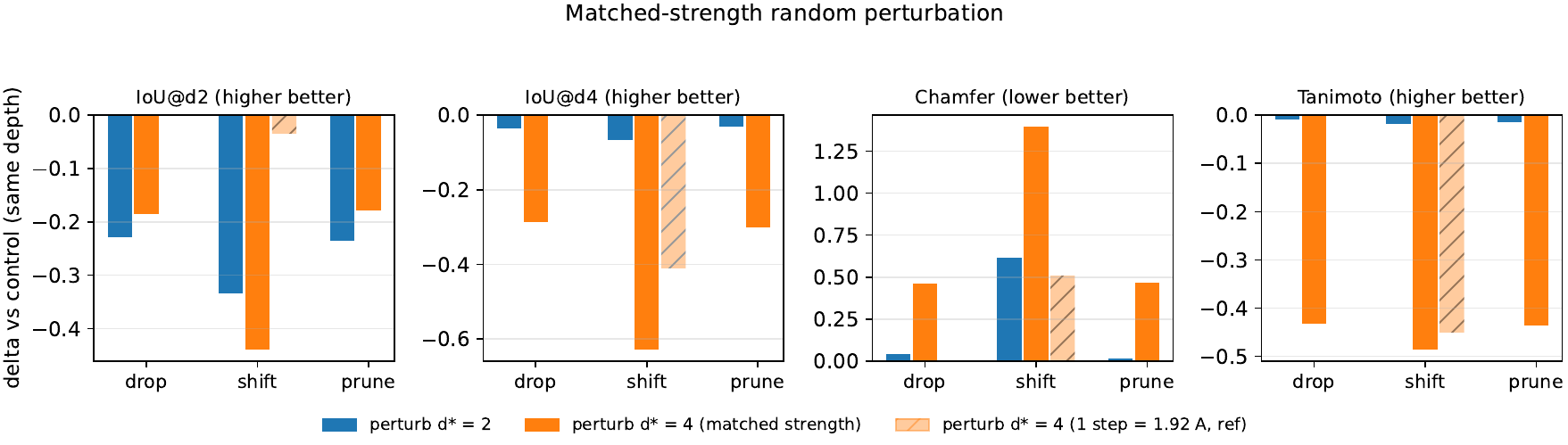}
  \caption{Matched-strength random perturbations over 128 references and 256
  generations per arm. Deletion and pruning remove the same depth-2-derived
  physical region at both depths, and matched shifts move cells by
  7.68\,\AA{}. The hatched depth-4 one-step shift is the unmatched
  1.92\,\AA{} reference. Bars report changes from same-depth controls.}
  \label{app:fig:depth_perturb_matched}
\end{figure}

We project 128 GEOM-DRUG references to depth $d^*\in\{2,4\}$ and compare
deletion, shift, and pruning perturbations before regenerating deeper levels
and running the atom flow. Each operator, control, and depth uses 256 samples
(two per reference). To avoid confounding depth with physical perturbation
strength, we shift depth-4 cells four grid steps and apply depth-2 deletion or
pruning regions to every depth-4 cell they
cover. Matched deletion and
pruning reduce depth-2 IoU similarly at both injection depths, while matched
depth-4 shifting is more destructive because many small cells move
independently. Finer conditioning nevertheless preserves more unperturbed
fine detail: after matched deletion, absolute depth-4 IoU is 0.68 when
conditioning at depth 4 and 0.24 when conditioning at depth 2
(\cref{app:fig:depth_perturb_matched}).

\subsection{Sampling Steps and Candidate Selection}
\label{app:sampling_selection}

We compare 50, 25, and 10 steps with fixed checkpoints and candidate budgets,
retaining 500 samples per setting. Timings use one A100 40GB. OctMole timing
covers both flows and excludes loading, sorting, file writing, and evaluation;
the Uni-3DAR row uses its native generation-only protocol.

\begin{table}[t]
  \centering
  \caption{Sampling-step pilot with 500 final samples per setting. Time is
  pure generation time per retained final molecule in milliseconds.}
  \label{app:tab:steps}
  \tablestyle{3pt}{1.12}
  \begin{tabular}{llrrrrr}
    \toprule
    Dataset & Setting & Time & Mol. Sta. & Valid & Connected & Unique \\
    \midrule
    QM9 & Ours 50 & 327.61 & 100.00 & 100.00 & 100.00 & 98.40 \\
        & Ours 25 & 184.85 & 99.80 & 99.80 & 99.80 & 99.40 \\
        & Ours 10 & 98.50 & 98.20 & 99.40 & 98.80 & 99.80 \\
        & Uni-3DAR & 218.90 & 100.00 & 100.00 & 100.00 & 99.60 \\
    \midrule
    GEOM-DRUG & Ours 50 & 342.82 & 87.80 & 98.00 & 96.20 & 100.00 \\
              & Ours 25 & 181.17 & 86.00 & 97.60 & 96.40 & 100.00 \\
              & Ours 10 & 83.56 & 71.20 & 95.40 & 90.20 & 100.00 \\
              & Uni-3DAR & 228.74 & 81.00 & 97.20 & 91.40 & 100.00 \\
    \bottomrule
  \end{tabular}
\end{table}

Twenty-five steps are $1.77\times$ faster on QM9 and $1.89\times$ faster on
GEOM-DRUG than 50 steps, with nearby scores in this pilot; ten steps degrade
stability and connectedness. Because this experiment has only 500 final
samples, it does not establish equal large-sample diversity, and the main
evaluation retains 50 steps.
\Cref{app:tab:steps} contains the complete results.

\begin{figure}[t]
  \centering
  \includegraphics[width=0.5\linewidth]{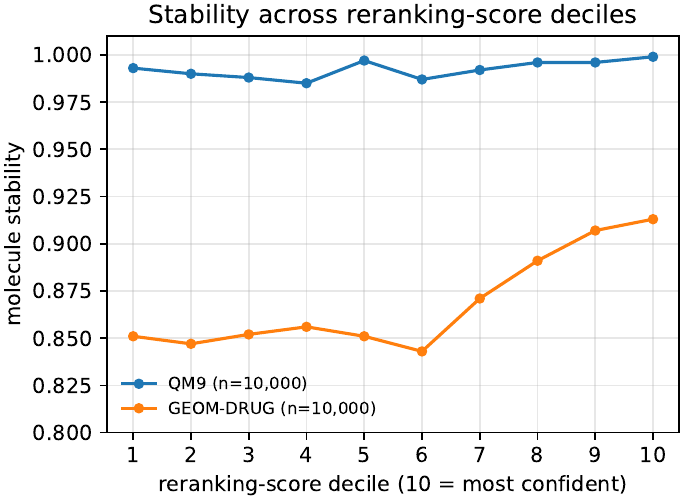}
  \caption{Molecule stability by reranking-score decile for 10,000 final
  samples per dataset. Higher-confidence bins generally improve stability on
  GEOM-DRUG, whereas QM9 is close to saturation throughout.}
  \label{app:fig:score_deciles}
\end{figure}

Ranking final samples by mean atom-type log probability generally separates
chemical quality on GEOM-DRUG: molecule stability rises from 85.1\% in the
lowest score decile to 91.3\% in the highest. QM9 is already saturated,
ranging from 98.5\% to 99.9\% across deciles
(\cref{app:fig:score_deciles}). On QM9, selection raises
training-set occupancy matches from approximately 13\% to 60--69\%, but an
occupancy match does not establish an exact chemical duplicate.

\section{Applications}
\label{app:sec:applications}
\subsection{Shape-Conditioned Ligand Reconstruction and Exploration}
\label{app:shape_application}

We use ligands from the test partition of the DiffDock time split of PDBBind
v2020~\citep{wang2004pdbbind,corso2023diffdock}. We evaluate all 354 ligands
that parse, contain a conformer, and fit the $30.72$\,\AA{} domain after
centering. We extract depth-4 or depth-6 occupancy and generate 20 candidates
per ligand with the GEOM-DRUG model, for 7,080 candidates per condition. No
atom types, local coordinates, or pocket atoms are supplied. The domain filter
biases the cohort toward compact ligands, and this is shape-conditioned
generation rather than structure-based drug design.

\begin{figure}[t]
  \centering
  \includegraphics[width=0.5\linewidth]{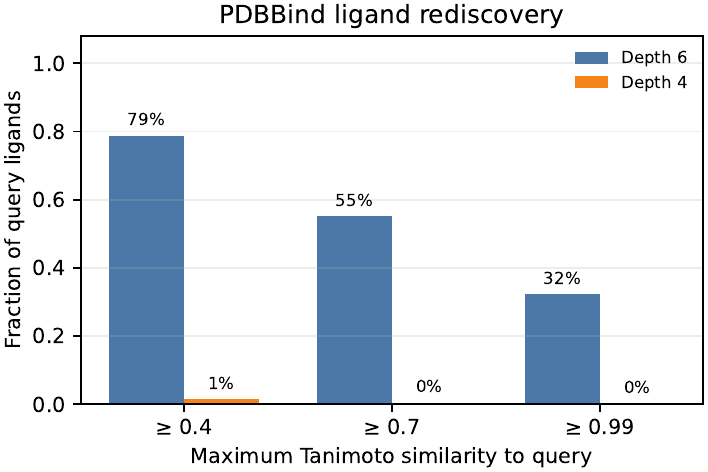}
  \caption{Ligand rediscovery from shape alone on 354 PDBBind ligands, with
  20 candidates per condition. Bars show the fraction of all query ligands
  whose best valid candidate reaches each Tanimoto threshold; unavailable
  query fingerprints count as failures.}
  \label{app:fig:pdbbind_rediscovery}
\end{figure}

At depth 6, the finest support is supplied and only atom types and local
features are sampled. For 32.2\% of ligands, at least one candidate reaches
Morgan-fingerprint Tanimoto similarity at least 0.99 to the query; 55.1\%
reach 0.7 and 78.8\% reach 0.4. At depth 4, 1.4\% reach 0.4 and none reach
0.99, demonstrating that coarse shape does not determine chemistry. These
ligand-level rates use all 354 queries as the denominator; 57 source SDFs do
not yield an evaluable query fingerprint and are therefore counted as
failures. \Cref{app:fig:pdbbind_rediscovery} summarizes these ligand-level
rediscovery rates.

\begin{figure}[t]
  \centering
  \includegraphics[width=0.7\linewidth]{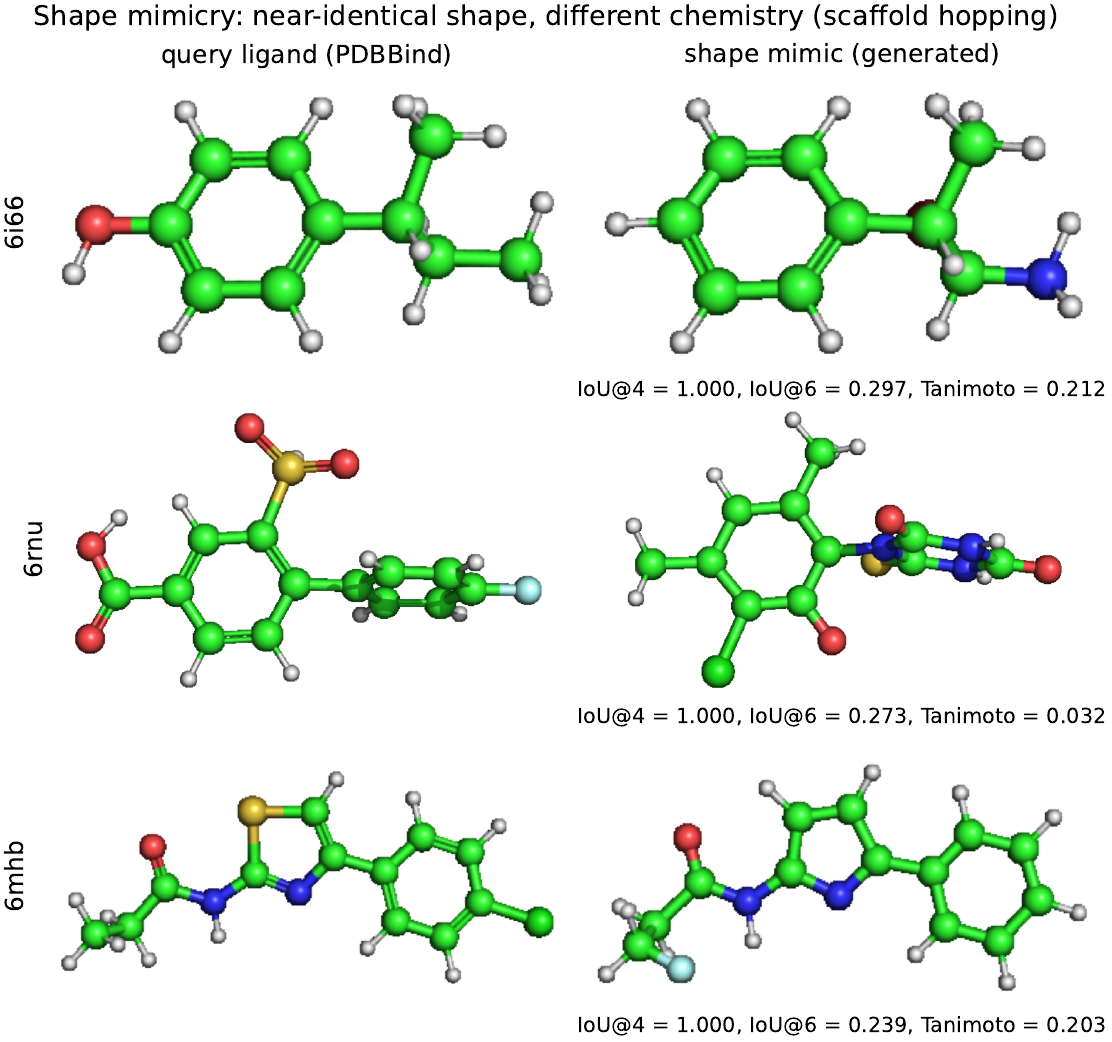}
  \caption{PDBBind queries (left) and generated depth-4 shape mimics (right)
  with a shared camera per row. Shapes are similar while elemental
  composition and inferred chemistry differ.}
  \label{app:fig:shape_mimicry_cases}
\end{figure}

At depth 4, generated molecules retain the query shape (mean occupancy IoU
0.909 and Chamfer distance 0.840\,\AA{}) while remaining chemically distinct
(mean Tanimoto 0.107 among evaluable pairs). Counting invalid samples and
unavailable fingerprints as failures, 69.6\% of all candidates are valid,
within 1\,\AA{} of the query shape, and have Tanimoto at most 0.3; 297 of 354
queries have at least one such candidate. Representative chemically distinct
shape mimics are shown in \cref{app:fig:shape_mimicry_cases}.

Coarse candidates have lower molecule stability than fine candidates, and
occupancy contains no pharmacophore or interaction information. Shape
retention alone therefore does not imply retained binding.

\subsection{Regional Fragment Editing}
\label{app:regional_editing}

We construct 9,943 candidates from distinct GEOM-DRUG validation molecules,
retain the finest occupied cells on one side of a median plane perpendicular
to the longest molecular axis, and regrow the other side from its coarse
depth-4 occupancy. Confidence reranking retains 2,000 final outputs. The
unconditional models are reused without editing-specific training. Retained
cells fix spatial support rather than atom identity, so the retained region is
a spatial scaffold rather than a chemically preserved fragment.

\begin{figure}[t]
  \centering
  \includegraphics[width=0.75\linewidth]{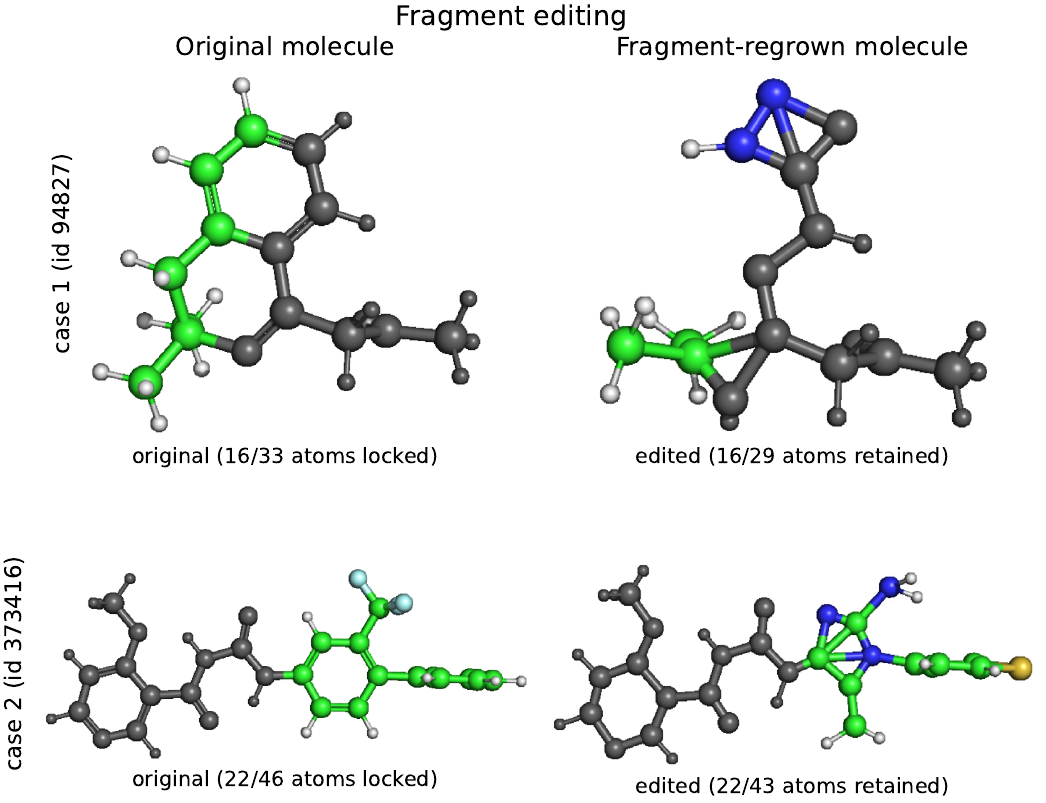}
  \caption{Regional fragment editing examples. Each row shows a different
  source molecule and its edited output. Dark gray marks atoms occupying the
  retained finest-level cells, while the colored region is regrown from its
  depth-4 support; atom identities and within-cell coordinates are generated
  throughout.}
  \label{app:fig:fragment_editing}
\end{figure}

Across the 2,000 edited outputs, validity is 81.9\%, connectedness is 46.8\%,
atom stability is 95.3\%, and molecule stability is 22.3\%. Merging retained
and regenerated supports creates an out-of-distribution seam that cell
corruption only partially addresses; examples are shown in
\cref{app:fig:fragment_editing}.

\subsection{Two-Fragment Linker Generation}
\label{app:linker_generation}

We take 100 GEOM-DRUG molecules with at least 18 heavy atoms and split each
along the heavy-atom bond-graph diameter into two end fragments of at least
five heavy atoms and a clean linear linker of at least two heavy atoms. End
fragment occupancy is locked at depth 6, while depth-4 occupancy in the linker
region provides a coarse corridor. Eight candidates are generated per target,
and the highest-confidence candidate is retained.

\begin{figure}[t]
  \centering
  \includegraphics[width=0.75\linewidth]{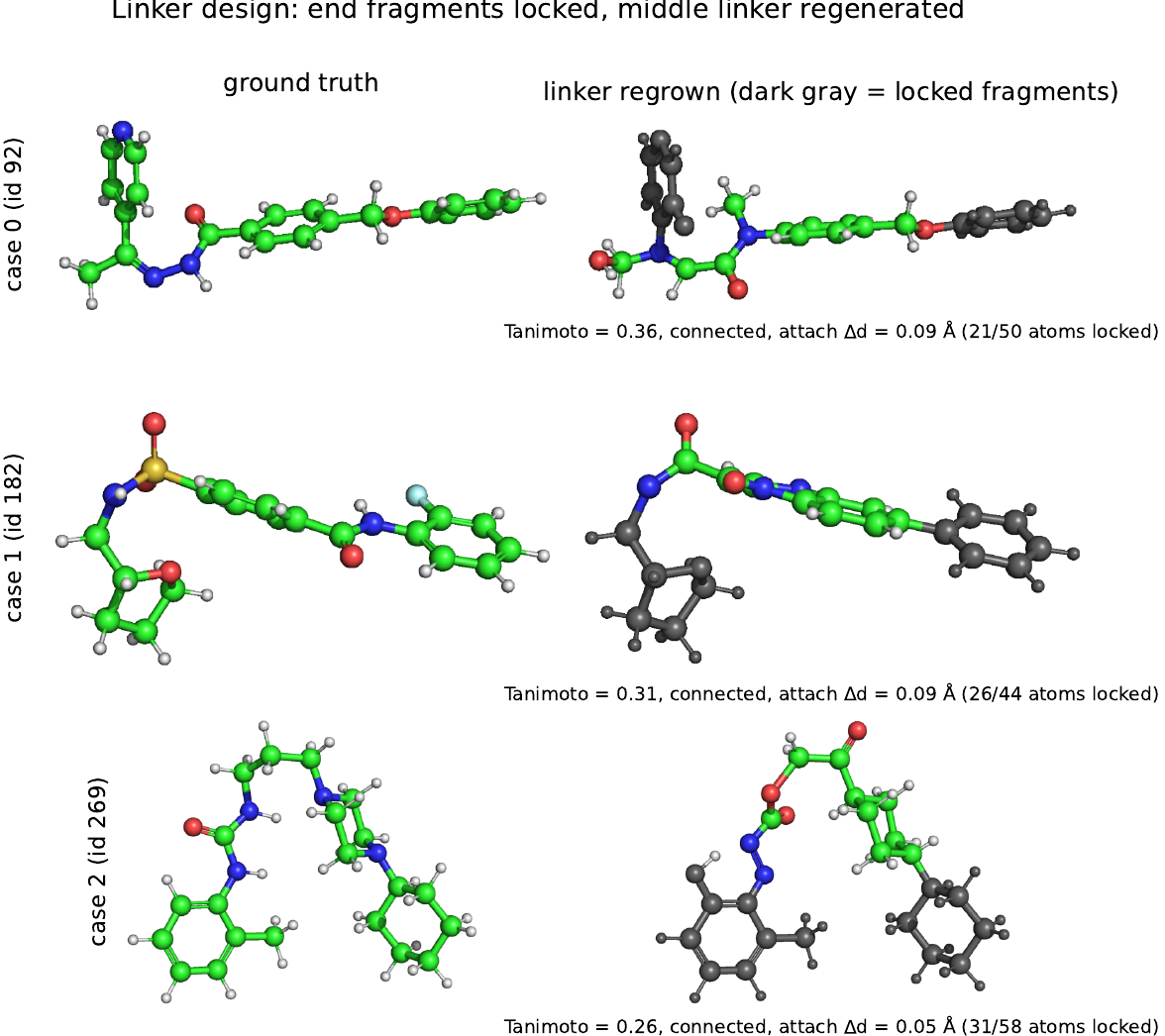}
  \caption{Three successful linker-generation cases. End-fragment occupancy
  is fixed in dark gray, and the regenerated linker is shown in element
  colors.}
  \label{app:fig:linker_cases}
\end{figure}

\begin{figure}[t]
  \centering
  \includegraphics[width=0.9\linewidth]{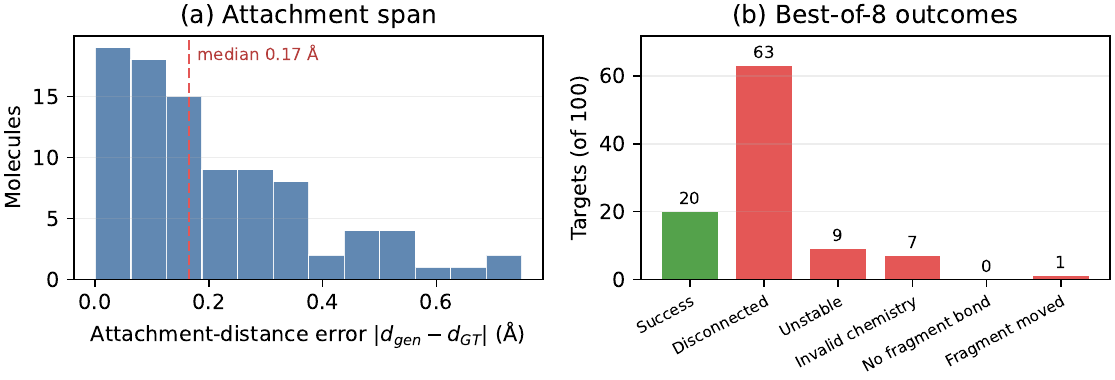}
  \caption{Linker-generation diagnostics on 100 targets. The left panel shows
  attachment-span errors for the 92 best-of-eight outputs with matched
  attachment atoms; the right panel classifies all selected outputs.
  Disconnection is the dominant outcome.}
  \label{app:fig:linker_stats}
\end{figure}

The median attachment-span error is 0.165\,\AA{} (mean 0.212\,\AA{}) over the
92 targets for which both attachment atoms can be matched. The best-of-eight
candidate is connected for 37.0\% of targets, compared with 25.6\% across all
800 candidates. Full success---connected, valid, stable, span error below
1\,\AA{}, and new bonds to both fragments---is achieved for 20 of 100 targets
(20\%). Disconnection is the dominant outcome for the selected candidates
(63 targets). Without a dedicated linker baseline, this is a capability
demonstration rather than a solved design task. Quantitative outcomes are
reported in \cref{app:tab:linker,app:fig:linker_stats}, with successful cases
in \cref{app:fig:linker_cases}.

\begin{table}[t]
  \centering
  \caption{Linker generation quality over 100 targets using best-of-eight
  candidates. Ground-truth molecules are re-evaluated under the same
  protocol.}
  \label{app:tab:linker}
  \tablestyle{6pt}{1.12}
  \begin{tabular}{lcccc}
    \toprule
    & Connected & Valid & Atom Sta. & Mol. Sta. \\
    \midrule
    Generated & 37.0 & 82.0 & 95.5 & 27.0 \\
    All candidates & 25.6 & 79.1 & 93.1 & 13.5 \\
    Ground truth & 100.0 & 94.0 & -- & 92.0 \\
    \bottomrule
  \end{tabular}
\end{table}

\subsection{Shape Interpolation}
\label{app:shape_interpolation}

We select 20 GEOM-DRUG reference pairs whose heavy-atom counts differ by at
most three and whose depth-4 occupancy IoU is 0.11--0.26. After centroid
alignment and a proper PCA rotation, we interpolate signed distance fields of
the two depth-4 occupancies at
$t\in\{0,0.25,0.5,0.75,1\}$. Generation then resumes from depth 4. Eight
candidates are sampled per pair and interpolation point, and the
highest-confidence candidate is retained.

\begin{figure}[t]
  \centering
  \includegraphics[width=\linewidth]{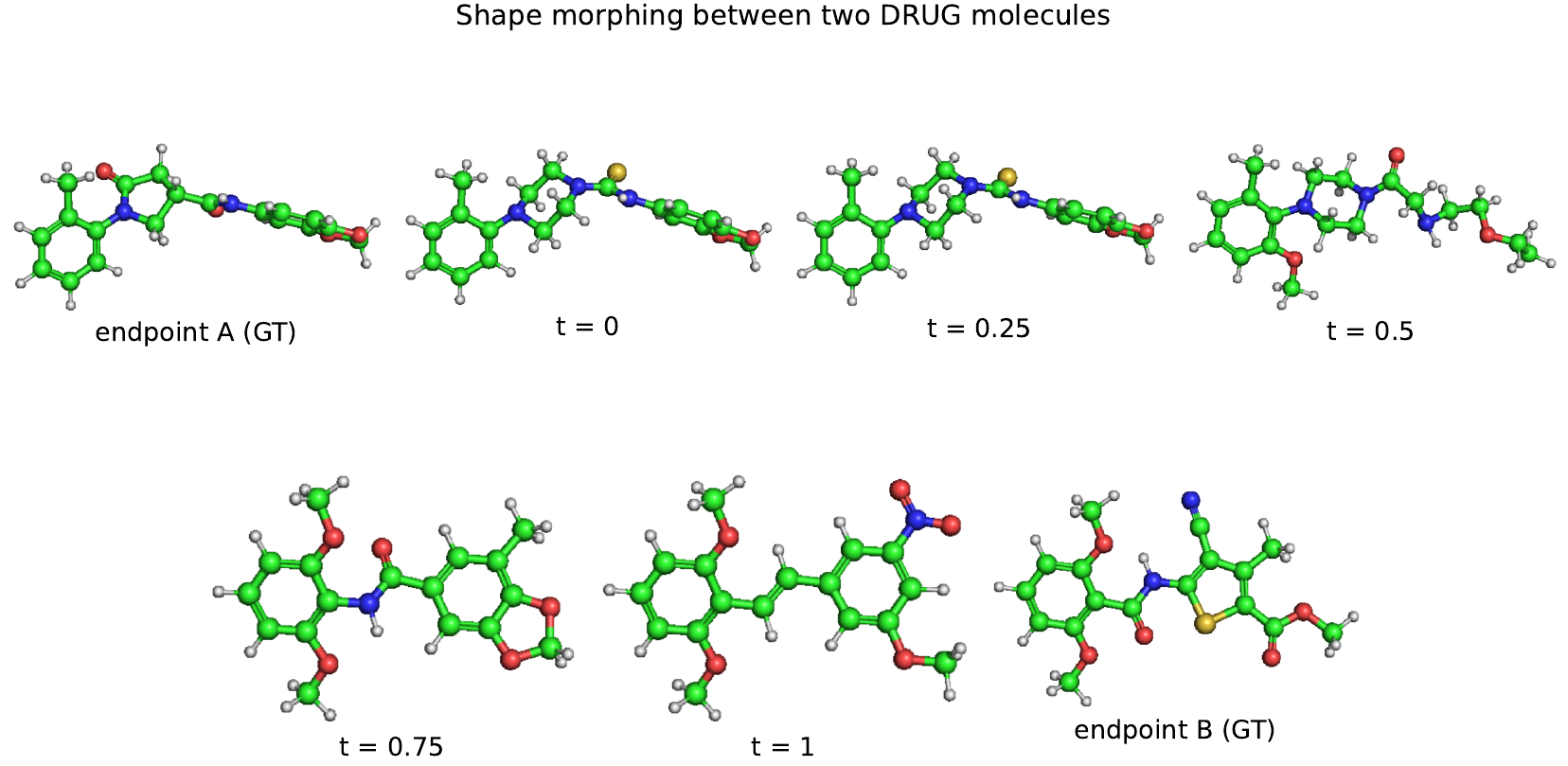}
  \caption{Shape interpolation for one representative GEOM-DRUG pair.
  Intermediate frames are best-of-eight generations from interpolated
  depth-4 occupancy.}
  \label{app:fig:morph_strip}
\end{figure}

\begin{figure}[t]
  \centering
  \includegraphics[width=0.5\linewidth]{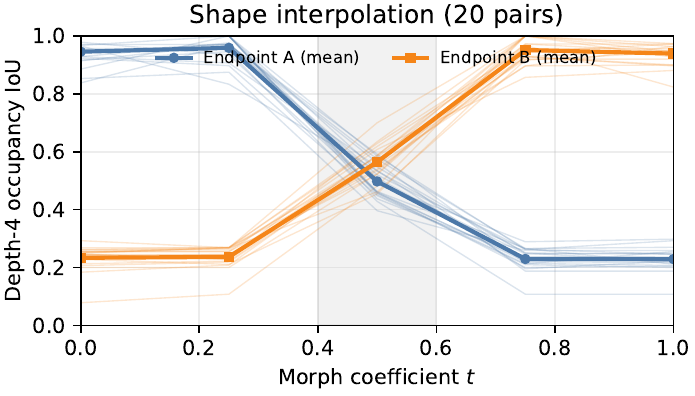}
  \caption{Depth-4 occupancy IoU to the two endpoints across 20 interpolation
  pairs. Thin curves show individual pairs and bold curves show their means;
  the mean curves cross near $t=0.5$, forming a geometric intermediate.}
  \label{app:fig:morph_iou}
\end{figure}

IoU to endpoint A decreases strongly with $t$, while IoU to endpoint B
increases, with mean per-pair correlations of $-0.936$ and 0.943,
respectively. The mean curves cross near $t=0.5$, where their IoUs are 0.50
and 0.56. Of the 60 selected intermediate outputs, 59 are valid and 53 are
molecule-stable. Mean endpoint IoU is 0.94--0.95 rather than one because the
occupancy flow may prune or complete individual supplied cells; finer depth-6
occupancy is regrown rather than copied.
\Cref{app:fig:morph_strip,app:fig:morph_iou} show a representative sequence
and the aggregate transition toward the two endpoints.

\end{document}